\documentclass[runningheads,]{llncs}

\usepackage{amsmath,amssymb} 
\usepackage{graphicx}
\usepackage{booktabs}
\usepackage{float}
\usepackage{algorithm2e}
\usepackage{xcolor}
\usepackage[width=122mm,left=12mm,paperwidth=146mm,height=193mm,top=12mm,paperheight=217mm]{geometry}

\usepackage{graphicx}
\usepackage{url} 

\usepackage[breaklinks]{hyperref}
\usepackage{calc}
\usepackage{microtype}
\usepackage{multirow}
\usepackage{textcomp}
\usepackage[export]{adjustbox}
\usepackage{array} 
\usepackage{tikz}
\usepackage{nth}

\makeatletter
\define@key{Gin}{xviewport}[]{%
  \begingroup\edef\x{%
    \endgroup\noexpand\setkeys{Gin}{viewport=\x@viewport}%
  }\x
}

\newlength{\Xoffset}
\newlength{\Yoffset}

\newcommand*{\setviewport}[4]{%
  \def\x@viewport{%
    {\the\dimexpr#1-\Xoffset} 
    {\the\dimexpr#2-\Yoffset}
    {\the\dimexpr#3-\Xoffset}
    {\the\dimexpr#4-\Yoffset}%
  }%
}
\makeatother

\makeatletter
\define@key{Gin}{expandoptions}{%
  \begingroup
  \edef\x{\endgroup
    \noexpand\setkeys{Gin}{#1}%
  }\x
}
\makeatother

\newcommand{\cstress}[1]{{\color{blue}{#1}}}
\definecolor{segPurple}{RGB}{64 0 128}
\definecolor{segRed}{RGB}{64 0 0}

\newlength{\tP}
\newlength{\twidth}
\newlength{\imgSize}
\newcommand{\Mod}[1]{\ (\textnormal{mod}\ #1)}

\newcommand{\IGNORE}[1]{}

\usetikzlibrary{shadows,calc}

\colorlet{innercolor}{black!60}
\colorlet{outercolor}{gray!05}

\usetikzlibrary{backgrounds}

\pgfdeclarelayer{shadow} 
\pgfsetlayers{shadow,main}

\begin{document}
\pagestyle{headings}
\mainmatter
\def\ECCV18SubNumber{3124}  

\title{Improving Shape Deformation in\\Unsupervised Image-to-Image Translation} 

\titlerunning{Improving Shape Deformation in Unsupervised Image-to-Image Translation}

\author{Aaron Gokaslan\inst{1}, Vivek Ramanujan\inst{1}, Daniel Ritchie\inst{1},\\Kwang In Kim\inst{2}, James Tompkin\inst{1}}
\index{Kim, Kwang In}

\authorrunning{Gokaslan et al.}

\institute{Brown University, USA \and University of Bath, UK}

\maketitle

\begin{abstract}
Unsupervised image-to-image translation techniques are able to map local texture between two domains, but they are typically unsuccessful when the domains require larger shape change. 
Inspired by semantic segmentation, we introduce a discriminator with dilated convolutions that is able to use information from across the entire image to train a more context-aware generator. 
This is coupled with a multi-scale perceptual loss that is better able to represent error in the underlying shape of objects. 
We demonstrate that this design is more capable of representing shape deformation in a challenging toy dataset, plus in complex mappings with significant dataset variation between humans, dolls, and anime faces, and between cats and dogs.

\keywords{Generative adversarial networks \and Image translation.} 
\end{abstract}

\section{Introduction}
Unsupervised image-to-image translation is the process of learning an arbitrary mapping between image domains without labels or pairings. 
This can be accomplished via deep learning with generative adversarial networks (GANs), through the use of a discriminator network to provide instance-\IGNORE{or mini-batch-}specific generator training, and the use of a cyclic loss to overcome the lack of supervised pairing. 
Prior works such as DiscoGAN~\cite{kim2017} and CycleGAN~\cite{zhuICCV2017} are able to transfer sophisticated local texture appearance between image domains, such as translating between paintings and photographs. 
However, these methods often have difficulty with objects that have both related appearance and shape changes; for instance, when translating between cats and dogs.

Coping with shape deformation in image translation tasks requires the ability to use spatial information from across the image. 
For instance, we cannot expect to transform a cat into a dog by simply changing the animals' local texture. 
From our experiments, networks with fully connected discriminators, such as DiscoGAN, are able to represent larger shape changes given sufficient network capacity, but train much slower~\cite{isolaCVPR2017} and have trouble resolving smaller details. 
Patch-based discriminators, as used in CycleGAN, work well at resolving high frequency information and train relatively quickly~\cite{isolaCVPR2017}, but have a limited `receptive field' for each patch that only allows the network to consider spatially local content. 
These networks reduce the amount of information received by the generator. 
Further, the functions used to maintain the cyclic loss prior in both networks retains high frequency information in the cyclic reconstruction, which is often detrimental to shape change tasks.

\begin{figure}[t]
    \centering
    \includegraphics[,,width=0.45\linewidth]{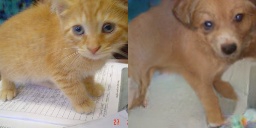}%
    \quad
    \includegraphics[,,width=0.45\linewidth]{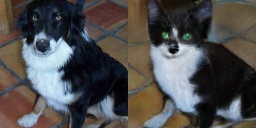}
    \caption{Our approach translates texture appearance and complex head and body shape changes between the cat and dog domains (left: input; right: translation).}
    \label{fig:teaser}
\end{figure}

We propose an image-to-image translation system, designated \emph{GANimorph}, to address shortcomings present in current techniques. 
To allow for patch-based discriminators to use more image context, we use dilated convolutions in our discriminator architecture \cite{yu2015multi}. 
This allows us to treat discrimination as a semantic segmentation problem: the discriminator outputs per-pixel real-vs.-fake decisions, each informed by global context.
This per-pixel discriminator output facilitates more fine-grained information flow from the discriminator to the generator.
We also use a multi-scale structure similarity perceptual reconstruction loss to help represent error over image areas rather than just over pixels.
We demonstrate that our approach is more successful on a challenging shape deformation toy dataset than previous approaches.
We also demonstrate example translations involving both appearance and shape variation by mapping human faces to dolls and anime characters, and mapping cats to dogs (Figure~\ref{fig:teaser}).

The source code to our GANimorph system and all datasets are online: \href{https://github.com/brownvc/ganimorph/}{https://github.com/brownvc/ganimorph/}.

\section{Related Work}
\paragraph{Image-to-image Translation.} Image analogies provides one of the earliest examples of image-to-image translation \cite{hertzmann2001image}. The approach relies on non-parametric texture synthesis and can handle transformations such as seasonal scene shifts~\cite{laffont2014transient}, color and texture transformation, and painterly style transfer. Despite the ability of the model to learn texture transfer, the model cannot affect the shape of objects. Recent research has extended the model to perform visual attribute transfer using neural networks \cite{liao2017,he2017neuralcolor}. However, despite these improvements, deep image analogies are unable to achieve shape deformation.

\paragraph{Neural Style Transfer.} These techniques show transfer of more complex artistic styles than image analogies \cite{gatysCVPR2015}. They combine the style of one image with the content of another by matching the Gram matrix statistics of early-layer feature maps from neural networks trained on general supervised image recognition tasks. Further, Duomiln et al.~\cite{dumoulin2016} extended Gatys et al.'s technique to allow for interpolation between pre-trained styles, and Huang et al.~\cite{huang2017arbitrarystyle} allowed real-time transfer. Despite this promise, these techniques have difficulty adapting to shape deformation, and empirical results have shown that these networks only capture low-level texture information \cite{bau2017}. Reference images can affect brush strokes, color palette, and local geometry, but larger changes such as anime-style combined appearance and shape transformations do not propagate.

\paragraph{Generative Adversarial Networks.} Generative adversarial networks (GANs) have produced promising results in image editing \cite{Liang2017}, image translation \cite{isolaCVPR2017}, and image synthesis \cite{goodfellow2014}. These networks learn an adversarial loss function to distinguish between real and generated samples. Isola et al.~\cite{isolaCVPR2017} demonstrated with Pix2Pix that GANs are capable of learning texture mappings between complex domains. However, this technique requires a large number of explicitly-paired samples. Some such datasets are naturally available, e.g., registered map and satellite photos, or image colorization tasks. We show in our supplemental material that our approach is also able to solve these limited-shape-change problems.
    
For specific domains such as faces, prior work has achieved domain transfer without explicit pairing. For instance, Taigman et al.~\cite{taigman2016emoji} tackled the problem of generating a personal emoji avatar from a photograph of a human face. Their technique requires a pre-trained facial attribute classifier, plus domain-specific and task-specific supervised labels for the photorealistic domain. Wolf et al.~\cite{wolf2017avatar} improved the generation by learning an underlying data generating avatar parameterization to create new avatars. Such a technique requires an existing and easily-parameterized model, and therefore cannot cope with more complex art styles, avatars, or avatar scenes, which are difficult to parameterize.

\paragraph{Unsupervised Image Translation GANs.} Pix2Pix-like architectures have been extended to work with unsupervised pairs \cite{kim2017,zhuICCV2017}. Given image domains X and Y, these approaches work by learning a cyclic mapping from X$\rightarrow$Y$\rightarrow$X and Y$\rightarrow$X$\rightarrow$Y. This creates a bijective mapping that prevents mode collapse in the unsupervised case. We build upon the DiscoGAN \cite{kim2017} and CycleGAN \cite{zhuICCV2017} architectures, which themselves extend Coupled GANs for style transfer \cite{liu2016}. We seek to overcome their shape change limitations through more efficient learning and expanded discriminator context via dilated convolutions, and by using a cyclic loss function that considers multi-scale frequency information (Table \ref{tab:dilatedVsDense}).

\begin{table}[t]
    \centering
    \makeatletter
    \define@key{Gin}{tSize}[true]{%
        \edef\@tempa{{Gin}{width=.23\linewidth, keepaspectratio}}%
        \expandafter\setkeys\@tempa
    }
    \makeatother
    \newcommand{\tableTitleSize}[1]{\footnotesize{#1}}
    \begin{minipage}[c]{0.69\linewidth}
        \begin{tabular}{c c c c}
            \tableTitleSize{Input} & \tableTitleSize{Patch based} & \tableTitleSize{Dense} & \tableTitleSize{Dilated} \\ 
            \includegraphics[tSize,frame,]{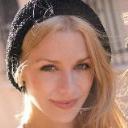}&
            \includegraphics[tSize,frame,]{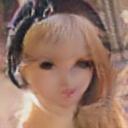} &
            \includegraphics[tSize,frame,]{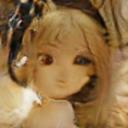}&
            \includegraphics[tSize,frame,, ]{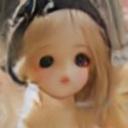} \\
            \includegraphics[tSize,frame,]{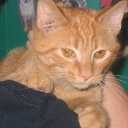}&
            \includegraphics[tSize,frame, ]{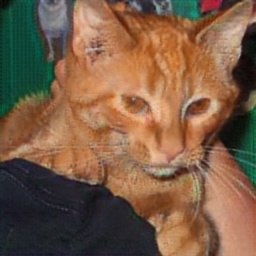}&
            \includegraphics[tSize,frame,, ]{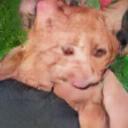}&
            \includegraphics[tSize,frame, ]{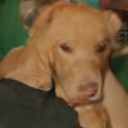}\\
        \end{tabular}
    \end{minipage}
    \begin{minipage}[c]{0.3\linewidth}
        \vspace{0.6cm} 
        \caption{\small Translating a human to a doll, and a cat to a dog. Dilated convolutions in the discriminator outperform both patch-based and dense convolution methods for image translations that require larger shape changes and small detail preservation.}
        \label{tab:dilatedVsDense}
    \end{minipage}
\end{table}

Other works tackle complementary problems. Yi et al.~\cite{yi2017dualgan} focus on improving high frequency features over CycleGAN in image translation tasks, such as texture transfer and segmentation. Shuang et al.~\cite{shuang2018dagan} examine adapting CycleGAN to wider variety in the domains---so-called instance-level translation. Liu et al.~\cite{liu2017} use two autoencoders to create a cyclic loss through a shared latent space with additional constraints. Several layers are shared between the two generators and an identity loss ensures that both domains resolve to the same latent vector. This produces some shape transformation in faces; however, the network does not improve the discriminator architecture to provide greater context awareness. 

One qualitatively different approach is to introduce object-level segmentation maps into the training set. Liang et al.'s ContrastGAN \cite{Liang2017} has demonstrated shape change by learning segmentation maps and combining multiple conditional cyclic generative adversarial networks. However, this additional input is often unavailable and time consuming to declare.

\section{Our Approach}

Crucial to the success of translation under shape deformation is the ability to maintain consistency over global shapes as well as local texture. Our algorithm adopts the cyclic image translation framework~\cite{kim2017,zhuICCV2017} and achieves the required consistency by incorporating a new dilated discriminator, a generator with residual blocks and skip connections, and a multi-scale perceptual cyclic loss.

\subsection{Dilated Discriminator}

Initial approaches used a global discriminator with a fully connected layer~\cite{kim2017}. Such a discriminator collapses an image to a single scalar value for determining image veracity. Later approaches~\cite{zhuICCV2017,Liang2017} used a patch-based DCGAN~\cite{radford2015dcgan} discriminator, initially developed for style transfer and texture synthesis~\cite{li2016precomputed}.
In this type of discriminator, each image patch is evaluated to determine a fake or real score. The patch-based approach allows for fast generator convergence by operating on each local patch independently.
This approach has proven effective for texture transfer, segmentation, and similar tasks.
However, this patch-based view limits the networks' awareness of global spatial information, which limits the generator's ability to perform coherent global shape change.

\paragraph{Reframing Discrimination as Semantic Segmentation.}
To solve this issue, we reframe the discrimination problem from determining real/fake images or subimages into the more general problem of finding real or fake regions of the image, i.e., a \emph{semantic segmentation} task.
Since the discriminator outputs a higher-resolution segmentation map, the information flow between the generator and discriminator increases.
This allows for faster convergence than using a fully connected discriminator, such as in DiscoGAN.

Current state-of-the-art networks for segmentation use dilated convolutions, and have been shown to require far fewer parameters than conventional convolutional networks to achieve similar levels of accuracy~\cite{yu2015multi}. 
Dilated convolutions provide advantages over both global and patch-based discriminator architectures. 
For the same parameter budget, they allow the prediction to incorporate data from a larger surrounding region. 
This increases the information flow between the generator and discriminator: by knowing that regions of the image contribute to making the image unrealistic, the generator can focus on that region of the image. 
An alternative way to think about dilated convolutions is that they allow the discriminator to implicitly learn context. 
While multi-scale discriminators have been shown to improve results and stability for high resolution image synthesis tasks~\cite{wang2017pix2pixHD}, we will show that incorporating information from farther away in the image is useful in translation tasks as the discriminator can determine where a region should fit into an image based on surrounding data.
For example, this increased spatial context helps localize the face of a dog relative to its body, which is difficult to learn from small patches or patches learned in isolation from their neighbors.
Figure~\ref{fig:architectureN} (right) illustrates our discriminator architecture.
\subsection{Generator}

\begin{figure}[t]
\includegraphics[width=\linewidth]{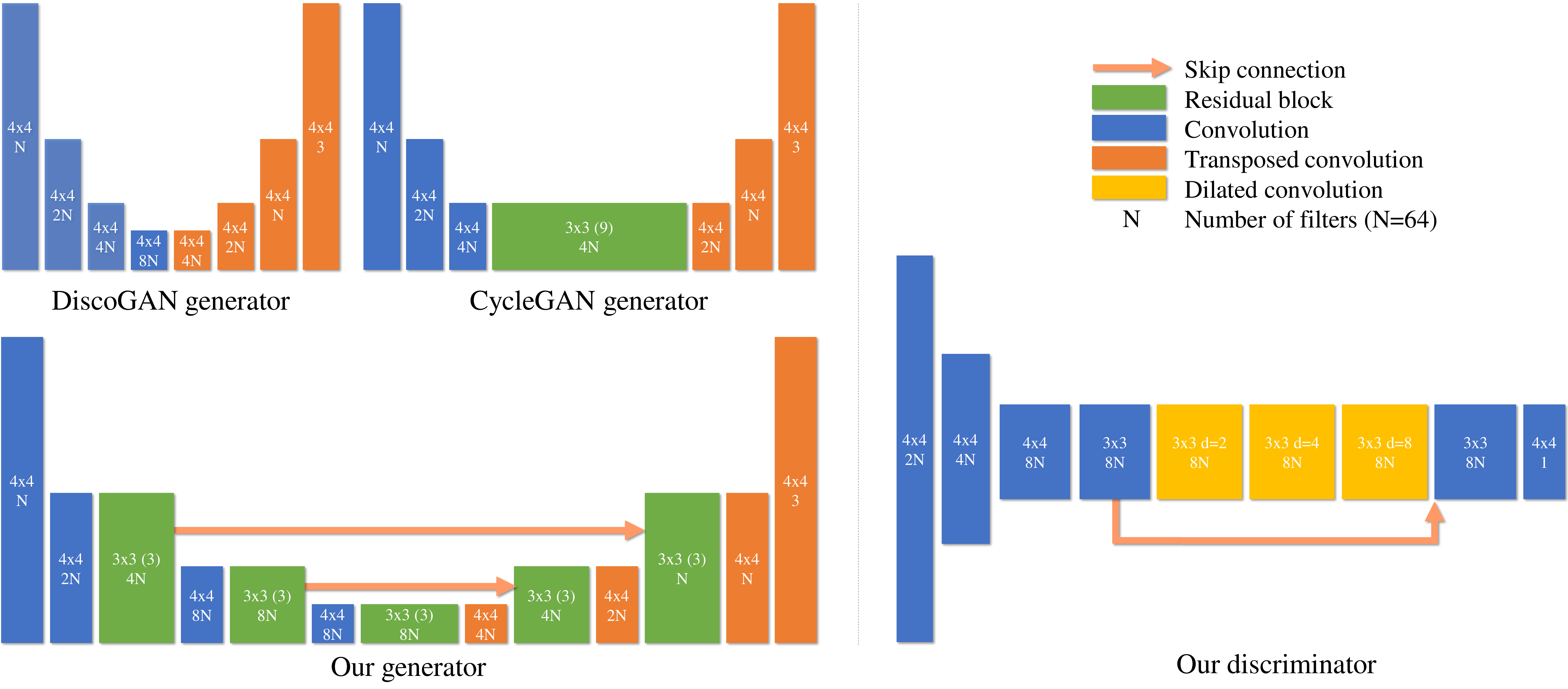}
\caption{\small (Left) Generators from different unsupervised image translation models. The skip connections and residual blocks are combined via concatenation as opposed to addition.
(Right) Our discriminator network architecture is a fully-convolutional segmentation network. Each colored block represents a convolution layer; block labels indicate filter size. In addition to global context from the dilations, the skip connection bypassing the dilated convolution blocks preserves the network's view of local context.}
\label{fig:architectureN}
\end{figure}

Our generator architecture builds on those of DiscoGAN and CycleGAN.
DiscoGAN uses a standard encoder-decoder architecture (Figure~\ref{fig:architectureN}, top left).
However, its narrow bottleneck layer can lead to output images that do not preserve all the important visual details from the input image. 
Furthermore, due to the low capacity of the network, the approach remains limited to low resolution images of size 64$\times$64. 
The CycleGAN architecture seeks to increase capacity over DiscoGAN by using a residual block to learn the image translation function~\cite{he2015resnet}. 
Residual blocks have been shown to work in extremely deep networks, and they are able to represent low frequency information~\cite{zeiler2014,bau2017}. 

However, using residual blocks at a single scale limits the information that can pass through the bottleneck and thus the functions that the network can learn. 
Our generator includes residual blocks at multiple layers of both the decoder and encoder, allowing the network to learn multi-scale transformations that work on both higher and lower spatial resolution features (Figure~\ref{fig:architectureN}, bottom left).

\subsection{Objective Function} 

\paragraph{Perceptual Cyclic Loss.}
As per prior unsupervised image-to-image translation work~\cite{kim2017,Liang2017,liu2017,zhuICCV2017,yi2017dualgan}, we use a cyclic loss to learn a bijective mapping between two image domains.
However, not all image translation functions can be perfectly bijective, e.g., when one domain has smaller appearance variation, like human face photos vs.~anime drawings. 
When all information in the input image cannot be preserved in the translation, the cyclic loss term should aim to preserve the most important information. Since the network should focus on image attributes of importance to human viewers, we should choose a perceptual loss that emphasizes shape and appearance similarity between the generated and target images.

Defining an explicit shape loss is difficult, as any explicit term requires known image correspondences between domains.
These do not exist for our examples and our unsupervised setting. 
Further, including a more-complex perceptual neural network into the loss calculation imparts a significant computational and memory overhead. 
While using pretrained image classification networks as a perceptual loss can speed up style transfer~\cite{johnson2016ploss}, these do not work on shape changes as the pretrained networks tend only to capture low-level texture information \cite{bau2017}.

Instead, we use multi-scale structure similarity loss (MS-SSIM)~\cite{wang2004}. This loss better preserves features visible to humans instead of noisy high frequency information. MS-SSIM can also better cope with shape change since it can recognize geometric differences through area statistics. 
However, MS-SSIM alone can ignore smaller details, and does not capture color similarity well.
Recent work has shown that mixing MS-SSIM with L1 or L2 losses is effective for super resolution and segmentation tasks~\cite{zhao2017Loss}. Thus, we also add a lightly-weighted L1 loss term, which helps increase the clarity of generated images.

\paragraph{Feature Matching Loss.}

To increase the stability of the model, our objective function uses a feature matching loss~\cite{salimans2016fm}:
\begin{equation}
    \mathcal{L}_{\textnormal{FM}}(G, D) = \frac{1}{n-1}\sum_{i=1}^{n-1} \lVert\mathbb{E}_{x \sim p_{\text{data}}}f_i(x) - \mathbb{E}_{z \sim p_{z}}f_i(G(z))\rVert_{2}^{2}.
    \label{eq:featurematchloss}
\end{equation}
Where $f_i\in D(x)$ represents the raw activation potentials of the $i^{th}$ layer of the discriminator $D$, and $n$ is the number of discriminator layers.
This term encourages fake and real samples to produce similar activations in the discriminator, and so encourages the generator to create images that look more similar to the target domain. We have found this loss term to prevent generator mode collapse, to which GANs are often susceptible~\cite{kim2017,salimans2016fm,wang2017pix2pixHD}.

\paragraph{Scheduled Loss Normalization (SLN).}
In a multi-part loss function, linear weights are often used to normalize the terms with respect to one another, with previous works often optimizing a single set of weights.
However, finding appropriately-balanced weights can prove difficult without ground truth. Further, often a single set of weights is inappropriate because the magnitude of the loss terms changes over the course of training. Instead, we create a procedure to periodically renormalize each loss term and so control their relative values. This lets the user intuitively provide weights that sum to 1 to balance the loss terms in the model, without having knowledge of how their magnitudes will change over training.

Let $\mathcal{L}$ be a loss function, and let $\mathcal{X}_n = \{x_t\}_{t=1}^{bn}$ be a sequence of $n$ batches of training inputs, each $b$ images large, such that $\mathcal{L}(x_t)$ is the training loss at iteration $t$.
We compute an exponentially-weighted moving average of the loss:
\begin{equation}
    \mathcal{L}_{\textnormal{moavg}}(\mathcal{L}, \mathcal{X}_n) = (1 - \beta)\sum_{x_t\in\mathcal{X}_n} \beta^{bn - t} \mathcal{L}(x_t)^2 
\end{equation}
where $\beta$ is the decay rate.
We can renormalize the loss function by dividing it by this moving average.
If we do this on every training iteration, however, the loss stays at its normalized average and no training progress is made.
Instead, we schedule the loss normalization:
\[
\textnormal{SLN}(\mathcal{L}, \mathcal{X}_n, s) = \begin{cases}
     \mathcal{L}(\mathcal{X}_n)/(\mathcal{L}_{\textnormal{moavg}}(\mathcal{L}, \mathcal{X}_n) + \epsilon) &\text{if } n\Mod{s} = 1\\
    \mathcal{L}(\mathcal{X}_n)&\text{otherwise}
\end{cases}
\]
Here, $s$ is the scheduling parameter such that we apply normalization every $s$ training iterations. For all experiments, we use $\beta = 0.99$, $\epsilon = 10^{-10}$, and $s=200$. 

One other normalization difference between CycleGAN/DiscoGAN and our approach is the use of instance normalization~\cite{huang2017arbitrarystyle} and batch normalization~\cite{ioffe2015batchnorm}, respectively. We found that batch normalization caused excessive over-fitting to the training data, and so we used instance normalization.

\paragraph{Final Objective.}
Our final objective comprises three loss normalized terms: a standard GAN loss, a feature matching loss, and two cyclic reconstruction losses. Given image domains $X$ and $Y$, let $G: X\to Y$ map from $X$ to $Y$ and $F:Y\to X$ map from $Y$ to $X$. $D_{X}$ and $D_{Y}$ denote discriminators for $G$ and $F$, respectively.

For GAN loss, we combine normal GAN loss terms from Goodfellow et al.~\cite{goodfellow2014}:
\begin{align}
    \mathcal{L}_{\textnormal{GAN}} &= \mathcal{L}_{\textnormal{GAN}_X}(F,D_{X},Y,X) + \mathcal{L}_{\textnormal{GAN}_Y}(G, D_{Y},X, Y)
\end{align}

For feature matching loss, we use Equation \ref{eq:featurematchloss} for each domain:
\begin{equation}
\mathcal{L}_{\textnormal{FM}} = \mathcal{L}_{\textnormal{FM}_X}(G, D_X) + \mathcal{L}_{\textnormal{FM}_Y}(F, D_Y)
\end{equation}

For the two cyclic reconstruction losses, we consider structural similarity~\cite{wang2004} and an $\mathbb{L}_1$ loss. Let $X'= F(G(X))$ and $Y'= G(F(Y))$ be the cyclically-reconstructed input images. Then:
\begin{align}
    \mathcal{L}_{\textnormal{SS}}=& (1-\text{MS-SSIM}(X', X)) + (1-\text{MS-SSIM}(Y', Y))\\ 
    \mathcal{L}_{\textnormal{L1}}=& \lVert X'-X\rVert_1 + \lVert Y'-Y\rVert_1
\end{align}
where we compute MS-SSIM without discorrelation.

Our total objective function with scheduled loss normalization (SLN) is:
\begin{align}
    \label{eqn:GGANtot}
    \mathcal{L}_{\textnormal{total}} =& \lambda_{\textnormal{GAN}} \textnormal{SLN}(\mathcal{L}_{\textnormal{GAN}}) + \lambda_{\textnormal{FM}} \textnormal{SLN}(\mathcal{L}_{\textnormal{FM}})  + \nonumber \\ & \lambda_{\textnormal{CYC}} \textnormal{SLN}(\lambda_{\textnormal{SS}}\mathcal{L}_{\textnormal{SS}} + \lambda_{\textnormal{L1}} \mathcal{L}_{\textnormal{L1}})
\end{align}
with $\lambda_{\textnormal{GAN}} + \lambda_{\textnormal{FM}} + \lambda_{\textnormal{CYC}} = 1$, $\lambda_{\textnormal{SS}} + \lambda_{\textnormal{L1}} = 1$, and all coefficients $\geq 0$. We set $\lambda_{\textnormal{GAN}}=0.49$, $\lambda_{\textnormal{FM}}=0.21$, and $\lambda_{\textnormal{CYC}}=0.3$, and $\lambda_{\textnormal{SS}}=0.7$ and $\lambda_{\textnormal{L1}}=0.3$. Empirically, these helped to reduce mode collapse and worked across all datasets.

\subsection{Training}

The network architecture both consumes and output 128$\times$128 images.
All models trained within 3.2 days on a single NVIDIA Titan X GPU with a batch size of 16. 
The number of generator updates per step varied between 1 and 2 for each dataset depending on the dataset difficulty. 
Each update of the generator used separate data than in the update of the discriminator.

We train for 50--400 epochs depending on the domain, with 1,000 batches per epoch. Overall, this resulted in 400,000 generator updates over the course of training for difficult datasets (e.g., cat to dog) and 200,000 generator updates for easier datasets (e.g., human to doll). We empirically define a dataset as hard or easy if it is difficult to generate images in the domain.

\paragraph{Data Augmentation.} To help mitigate dataset overfitting, the following image augmentations were applied to each dataset: rescale to 1.1 input size, random horizontal flipping of the image, random rotation of up to 30 degrees in either direction, random rescaling, and random cropping of the image.

\section{Experiments}

\begin{figure}[t]
    \centering
    \begin{minipage}[c]{0.6\linewidth}
        \includegraphics[frame,width=0.45\linewidth]{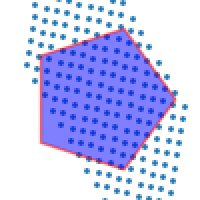}
        \includegraphics[frame,width=0.45\linewidth]{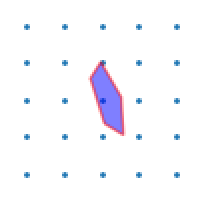}
    \end{minipage}\hfill
    \begin{minipage}[c]{0.4\linewidth}
        \vspace{-0.35cm}
        \caption{Toy Dataset (128$\times$128). \emph{Left:} $\mathcal{X}$ instance; a regular polygon with deformed dot matrix overlay. \emph{Right:} $\mathcal{Y}$ instance; a deformed polygon and dot lattice. The dot lattice provides information from across the image to the true deformation.} 
    \label{fig:toysetup}
    \end{minipage}
\end{figure}

\subsection{Toy Problem: Learning 2D Dot and Polygon Deformations}
We created a challenging toy problem to evaluate the ability of our network design to learn shape- and texture-consistent deformation. We define two domains: the regular polygon domain $X$ and its deformed equivalent $Y$ (Figure~\ref{fig:toysetup}). Each example $X_{s, h, d}\in X$ contains a centered regular polygon with $s\in \{3\ldots 7\}$ sides, plus a deformed matrix of dots overlaid. The dot matrix is computed by taking a unit dot grid and transforming it via $h$, a Gaussian random normal 2$\times$2 matrix, and a displacement vector $d$, a Gaussian normal vector in $\mathbb{R}^2$. The corresponding domain equivalent in $Y$ is $Y_{s, h, d}$, with instead the polygon transformed by $h$ and the dot matrix remaining regular. This construction forms a bijection from $X$ to $Y$, and so the translation problem is well-posed.

Learning a mapping from $X$ to $Y$ requires the network to use the large-scale cues present in the dot matrix to successfully deform the polygon, as local patches with a fixed image location cannot overcome the added displacement $d$. Table \ref{tab:toyresults} shows that DiscoGAN is unable to learn to map between either domain, and produces an output that is close to the mean of the dataset (off-white). CycleGAN is able to learn only local deformation, which produces hue shifts towards the blue of the polygon when mapping from regular to deformed spaces, and which in most cases produces an undeformed dot matrix when mapping from deformed to regular spaces. In contrast, our approach is significantly more successful at learning the deformation as the dilated discriminator is able to incorporate information from across the image.

\begin{table}[t]
    \setlength{\tabcolsep}{0.01pt}
    \renewcommand{\arraystretch}{0.05}%
    \providecommand\aport{}
    \newcommand{\tableTitleSize}[1]{\tiny{#1}}
    \newcommand{\tPix}{128}
    \setlength{\tP}{128\Xoffset}
        
    \makeatletter
    \define@key{Gin}{tSize}[true]{%
        \edef\@tempa{{Gin}{width=.110\linewidth, keepaspectratio}}%
        \expandafter\setkeys\@tempa
    }
    \makeatother

    \caption{Toy Dataset. When trying to estimate complex deformation, DiscoGAN collapses to the mean value of dataset (all white). CycleGAN is able to approximate the deformation of the polygon but not the dot lattice (right-hand side). Our approach is able to learn both under strong deformation.}
    \vspace{0.25cm}
    \label{tab:toyresults}
    \centering
    \rotatebox[origin=c]{90}{Regular to Deformed}
    \begin{tabular}[c]{c c c c}
        \tableTitleSize{Input} & \tableTitleSize{CycleGAN} & \tableTitleSize{DiscoGAN} & \tableTitleSize{Ours}\\
        \includegraphics[tSize,frame]{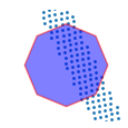} &
        \includegraphics[tSize,frame]{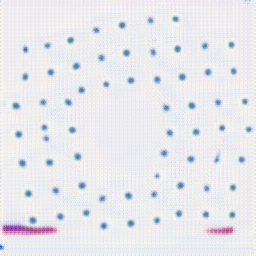} &
        \includegraphics[tSize,frame]{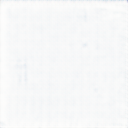} &
        \includegraphics[tSize,frame]{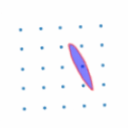} \\
        
        \includegraphics[tSize,frame]{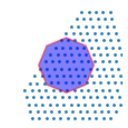} &
        \includegraphics[tSize,frame]{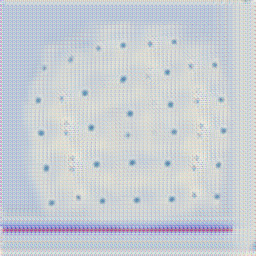} &
        \includegraphics[tSize,frame]{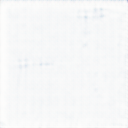} &
        \includegraphics[tSize,frame]{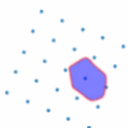} \\
        
        \includegraphics[tSize,frame]{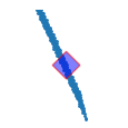} &
        \includegraphics[tSize,frame]{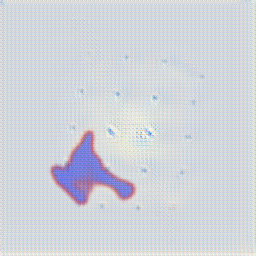} &
        \includegraphics[tSize,frame]{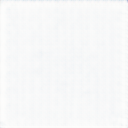} &
        \includegraphics[tSize,frame]{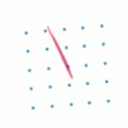} \\
        
        \includegraphics[tSize,frame]{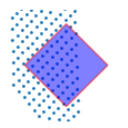} &
        \includegraphics[tSize,frame]{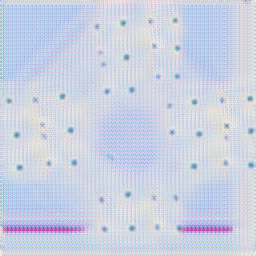} &
        \includegraphics[tSize,frame]{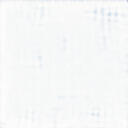} &        
        \includegraphics[tSize,frame]{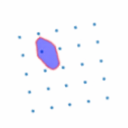} \\
    \end{tabular}
    \begin{tabular}[c]{c c c c}

        \tableTitleSize{Input} & \tableTitleSize{CycleGAN} & \tableTitleSize{DiscoGAN} & \tableTitleSize{Ours}  \\

        \includegraphics[tSize,frame]{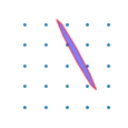} &
        \includegraphics[tSize,frame]{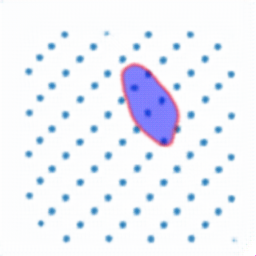} &
        \includegraphics[tSize,frame]{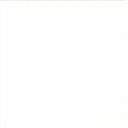} &
        \includegraphics[tSize,frame]{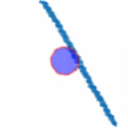} \\
        
        \includegraphics[tSize,frame]{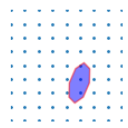} &
        \includegraphics[tSize,frame]{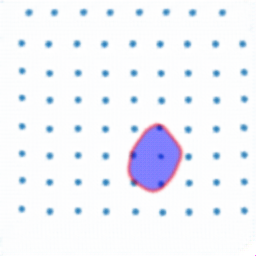} &
        \includegraphics[tSize,frame]{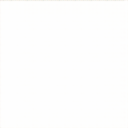} &
        \includegraphics[tSize,frame]{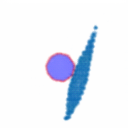} \\
        
        \includegraphics[tSize,frame]{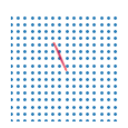} &
        \includegraphics[tSize,frame]{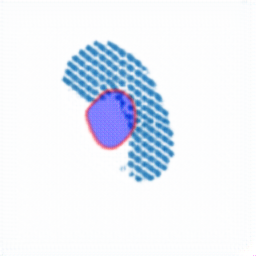} &
        \includegraphics[tSize,frame]{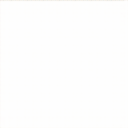} &
        \includegraphics[tSize,frame]{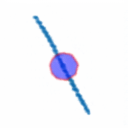} \\
        
        \includegraphics[tSize,frame]{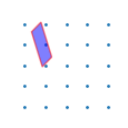} &
        \includegraphics[tSize,frame]{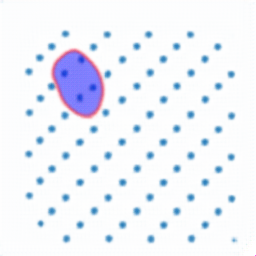} &
        \includegraphics[tSize,frame]{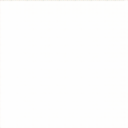} &
        \includegraphics[tSize,frame]{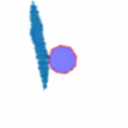} \\
    \end{tabular}
    \rotatebox[origin=c]{-90}{Deformed to Regular}\\
    \label{fig:toyresults}
\end{table}

\paragraph{Quantitative Comparison.}

As our output is a highly-deformed image, we estimate the learned transform parameters by sampling. We compute a Hausdorff distance between 500 point samples on the ground truth polygon and on the image of the generated polygon after translation: for finite sets of points $X$ and $Y$, $d(X, Y) = \max_{y\in Y}\min_{x\in X} \lVert x - y\rVert$. We hand annotate 220 generated polygon boundaries for our network, sampled uniformly at random along the boundary. Samples exist in a unit square with bottom left corner at (0, 0).

First, DiscoGAN fails to generate polygons at all, despite being able to reconstruct the original image. 
Second, for `regular to deformed', CycleGAN fails to produce a polygon, whereas our approach produces average Hausdorff distance of $0.20\pm 0.01$.
Third, for `deformed to regular', CycleGAN produces a polygon with distance of $0.21\pm 0.04$, whereas our approach has distance of $0.10\pm 0.03$. 
In the true dataset, note that regular polygons are centered, but CycleGAN only constructs polygons at the position of the original distorted polygon. Our network constructs a regular polygon at the center of the image as desired.

\subsection{Real-world Datasets}
\begin{figure}[t]
    \setlength{\imgSize}{.15\textwidth}

    \centering
    \includegraphics[width=\imgSize,frame,]{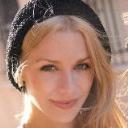}
    \includegraphics[width=\imgSize,frame,]{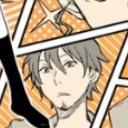}
    \includegraphics[width=\imgSize,frame,]{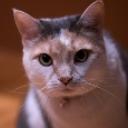}
    \includegraphics[width=\imgSize,frame,]{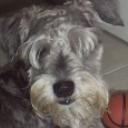}
    \includegraphics[width=\imgSize,frame,]{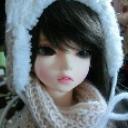}
    \includegraphics[width=\imgSize,, ]{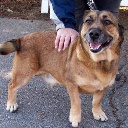}
    \caption{Face dataset examples, left to right: CelebA, Danbooru, Flickr Cat, Columbia Dog, Flickr Dolls, and Pets in the Wild.} 
    \label{fig:Dataset Ex}
\end{figure}

We evaluate our GANimorph system by learning mappings between several image datasets (Figure~\ref{fig:Dataset Ex}). For human faces, we use the aligned version of the CelebFaces Attribute dataset~\cite{liu2015faceattributes}, which contains 202,599 images.

\paragraph{Anime Faces.}
Previous works have noted that anime images are challenging to use with existing style transfer methods, since translating between a photoreal face and an anime-style face involves both shape and appearance variation. To test on anime faces, we create a large 966,777 image anime dataset crowdsourced from Danbooru~\cite{danbooru2017}. The Danbooru dataset has a wide variety of styles from super-deformed chibi-style faces, to realistically-proportioned faces, to rough sketches. Since traditional face detectors yield poor results on drawn datasets, we ran the Animeface filter~\cite{nagadomiAnimeface} on both datasets.

When translating humans to anime, we see an improvement in our approach for head pose and accessories such as glasses (Table~\ref{tab:mainresults}, \nth{3} row, right), plus a larger degree of shape deformation such as reduced face vertical height. 
The final line of each group represents a particularly challenging example.

\begin{table}[t]
    \setlength{\tabcolsep}{0.05em}
    \renewcommand{\arraystretch}{0.2}%
    \providecommand\aport{}
    \newcommand{\tableTitleSize}[1]{\tiny{#1}}
    \newcommand{\tPix}{128}
    \setlength{\tP}{128\Xoffset}
        
    \makeatletter
    \define@key{Gin}{tSize}[true]{%
        \edef\@tempa{{Gin}{width=.110\linewidth, keepaspectratio}}%
        \expandafter\setkeys\@tempa
    }
    \makeatother

    \caption{GANimorph can translate shape and style changes while retaining many input attributes such as hair color, pose, glasses, headgear, and background. CycleGAN and DiscoGAN are less successful both with just shape (human to doll) and with shape and style changes (human to anime).}
    \label{tab:mainresults}
    \centering
    \begin{tabular}[c]{c c}
    \rotatebox[origin=c]{90}{Photoreal to Anime}
    \begin{tabular}[c]{c c c c}
        \tableTitleSize{Input} & \tableTitleSize{CycleGAN} & \tableTitleSize{DiscoGAN} & \tableTitleSize{Ours}  \\

        \includegraphics[tSize,]{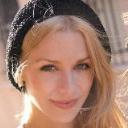} &
        \includegraphics[tSize,]{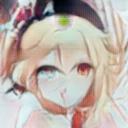} &
        \includegraphics[tSize,]{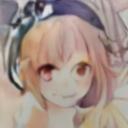} &
        \includegraphics[tSize,]{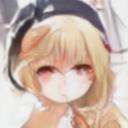} \\
        
        \includegraphics[tSize,]{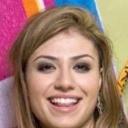} &
        \includegraphics[tSize,]{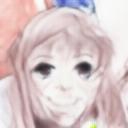} &
        \includegraphics[tSize,]{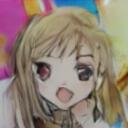} &
        \includegraphics[tSize,, ]{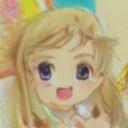} \\

        \includegraphics[tSize,]{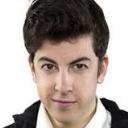} &
        \includegraphics[tSize,]{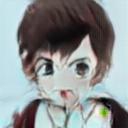} &
        \includegraphics[tSize,]{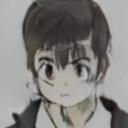} &
        \includegraphics[tSize,]{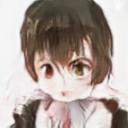} \\

        \includegraphics[tSize,]{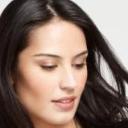} &
        \includegraphics[tSize,]{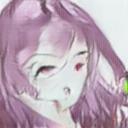} &
        \includegraphics[tSize,]{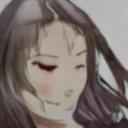} &
        \includegraphics[tSize,]{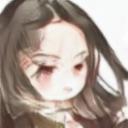} \\
    \end{tabular} &
    \hspace{0.4\tabcolsep}
    \begin{tabular}[c]{c c c c}
        \tableTitleSize{Input} & \tableTitleSize{CycleGAN} & \tableTitleSize{DiscoGAN} & \tableTitleSize{Ours}  \\

        \includegraphics[tSize,]{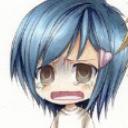} &
        \includegraphics[tSize,]{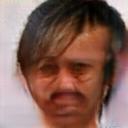} &
        \includegraphics[tSize,]{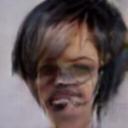} &
        \includegraphics[tSize,, ]{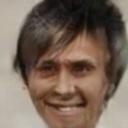} \\
        
        \includegraphics[tSize,]{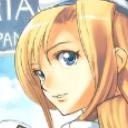} &
        \includegraphics[tSize,]{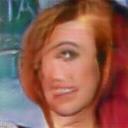} &
        \includegraphics[tSize,]{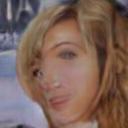} &
        \includegraphics[tSize,]{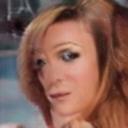} \\
        
        \includegraphics[tSize,]{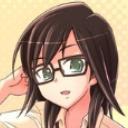} &
        \includegraphics[tSize,]{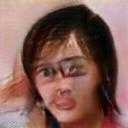} &
        \includegraphics[tSize,]{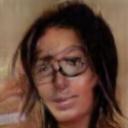} &
        \includegraphics[tSize,]{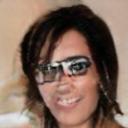} \\
        
        \includegraphics[tSize]{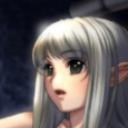} &
        \includegraphics[tSize]{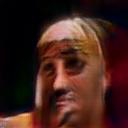} &
        \includegraphics[tSize]{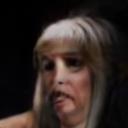} &
        \includegraphics[tSize]{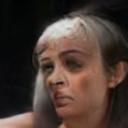} \\
    \end{tabular}
    \rotatebox[origin=c]{-90}{Anime to Photoreal}\\
    \vspace{2.5\tabcolsep} \\
    \rotatebox[origin=t]{90}{Human to Doll Face}
    \begin{tabular}[c]{c c c c}
        \centering
        \includegraphics[tSize,]{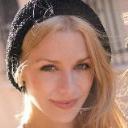} &
        \includegraphics[tSize,]{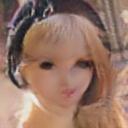} &
        \includegraphics[tSize,]{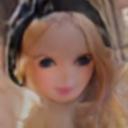} &
        \includegraphics[tSize,, ]{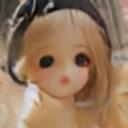} \\
        
        \includegraphics[tSize,]{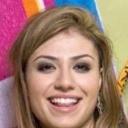} &
        \includegraphics[tSize,]{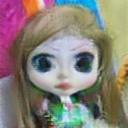} &
        \includegraphics[tSize,]{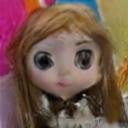} &
        \includegraphics[tSize,, ]{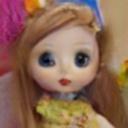} \\
        
        \includegraphics[tSize,]{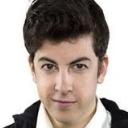} &
        \includegraphics[tSize,]{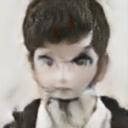} &
        \includegraphics[tSize,]{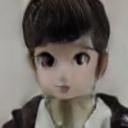} &
        \includegraphics[tSize,, ]{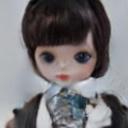} \\

        \includegraphics[tSize,]{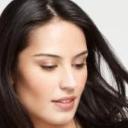} &
        \includegraphics[tSize,]{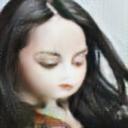} &
        \includegraphics[tSize,]{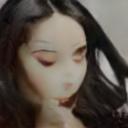} &
        \includegraphics[tSize,, ]{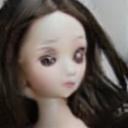} \\
        
        \tableTitleSize{Input} & \tableTitleSize{CycleGAN} & \tableTitleSize{DiscoGAN} & \tableTitleSize{Ours}  
    \end{tabular} &
    \hspace{0.4\tabcolsep}
    \begin{tabular}[c]{c c c c}
 
        \includegraphics[tSize,]{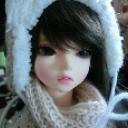} &
        \includegraphics[tSize,]{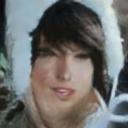} &
        \includegraphics[tSize,]{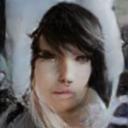} &
        \includegraphics[tSize,, ]{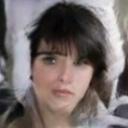} \\
        
        \includegraphics[tSize,]{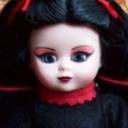} &
        \includegraphics[tSize,]{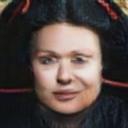} &
        \includegraphics[tSize,]{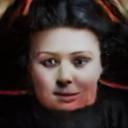} &
        \includegraphics[tSize,, ]{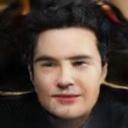} \\
        
        \includegraphics[tSize,]{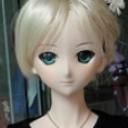} &
        \includegraphics[tSize,]{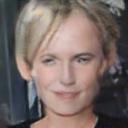} &
        \includegraphics[tSize,]{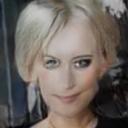} &
        \includegraphics[tSize,, ]{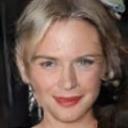} \\
        
        \includegraphics[tSize,]{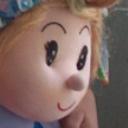} &
        \includegraphics[tSize,]{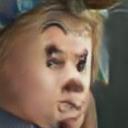} &
        \includegraphics[tSize,]{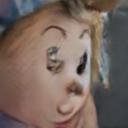} &
        \includegraphics[tSize,]{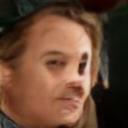} \\
        
        \tableTitleSize{Input} & \tableTitleSize{CycleGAN} & \tableTitleSize{DiscoGAN} & \tableTitleSize{Ours}
    \end{tabular}
    \rotatebox[origin=c]{-90}{Doll Face to Human}
    \end{tabular}
    \label{fig:results}
    \vspace{-0.5cm}
\end{table}

\paragraph{Doll Faces.}
To demonstrate that our algorithm can handle shape deformations with similar photographic appearance, we translate between the two domains of doll and human face photographs. Similar to Morsita et al.~\cite{morshita2016dolls}, we extracted 13,336 images from the Flickr100m dataset~\cite{ni2015flickr100m} using specific doll manufacturers as keywords. Then, we extract local binary patterns~\cite{ojala1994lbp} using OpenCV~\cite{opencv_library}, and use the Animeface filter for facial alignment~\cite{nagadomiAnimeface}. Stylizing human faces as dolls provides an informative test case: both domains have similar photorealistic appearance, so the translation task focuses on shape more than texture.

Table \ref{tab:mainresults}, bottom, shows that our architecture handles local deformation and global shape change better than CycleGAN and DiscoGAN, while preserving local texture similarity to the target domain. 
The second to last row on the right hand side shows that, with other networks, either the shape is malformed (DiscoGAN), or the shape shows artifacts from the original image or unnatural skin texture (CycleGAN). 
Our method matches skintones from the CelebA dataset, while capturing the overall facial structure and hair color of the doll. 
For a more difficult doll to human example in the bottom right-hand corner, while our transformation is not realistic, our method still creates more shape change than existing networks.

\paragraph{Pet Faces.}
To test whether our network could translate between animal domains, we constructed a dataset of 47,906 cat faces from Flickr100m\cite{ni2015flickr100m} dataset using OpenCV's~\cite{opencv_library} Haar cascade cat face detector.
This detector produces false positives, and occasionally detects a human face as a cat face.
We also use the 8,223-image Columbia Dog dataset~\cite{liu2012dog}, which comes with curated bounding boxes around the dog faces, which reduced the number of noisy results. Translation results are shown in Table \ref{fig:petfaces}.

\begin{table}[t]
        
    \makeatletter
    \define@key{Gin}{tSize}[true]{%
        \edef\@tempa{{Gin}{width=.117\linewidth, keepaspectratio}}%
        \expandafter\setkeys\@tempa
    }
    \makeatother
    \caption{Pet Faces: GANimorph can map poses across large variations in appearance, and does not incorrectly replace local texture without replacing the surrounding context.}
    \label{fig:petfaces}
    \setlength{\tabcolsep}{0.00em}
    \renewcommand{\arraystretch}{0.01}%
    \providecommand\aport{}
    \newcommand{\tableTitleSize}[1]{\tiny{#1}}
    
    \centering
   \begin{tabular}[c]{c c c c}
        \tableTitleSize{Input} & \tableTitleSize{CycleGAN} & \tableTitleSize{DiscoGAN} & \tableTitleSize{Ours} \\

        \includegraphics[tSize,frame,, ]{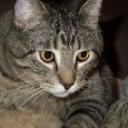} &
        \includegraphics[tSize,frame,, ]{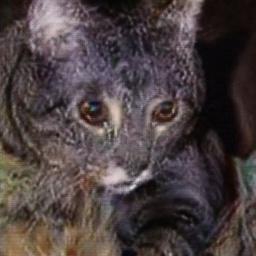} &
        \includegraphics[tSize,frame,, ]{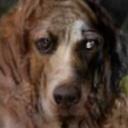} &
        \includegraphics[tSize,frame,, ]{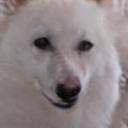}\\

        \includegraphics[tSize,frame,, ]{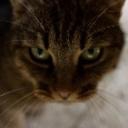} &
        \includegraphics[tSize,frame,, ]{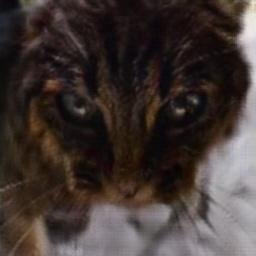} &
        \includegraphics[tSize,frame,, ]{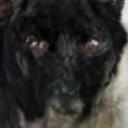} &
        \includegraphics[tSize,frame,, ]{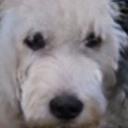}\\

        \includegraphics[tSize,frame,, ]{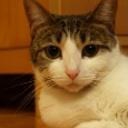} &
        \includegraphics[tSize,frame,, ]{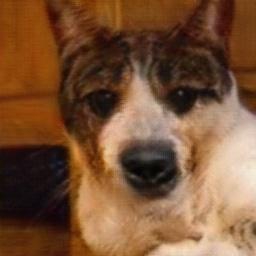} &
        \includegraphics[tSize,frame,, ]{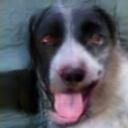} &
        \includegraphics[tSize,frame,, ]{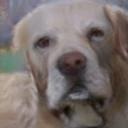}\\
        \tableTitleSize{Input} & \tableTitleSize{CycleGAN} & \tableTitleSize{DiscoGAN} & \tableTitleSize{Ours} \\

\end{tabular} %
\begin{tabular}{c c c c}
        \tableTitleSize{Input} & \tableTitleSize{CycleGAN} & \tableTitleSize{DiscoGAN} & \tableTitleSize{Ours} \\

        \includegraphics[tSize,frame,, ]{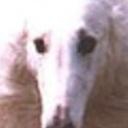} &
        \includegraphics[tSize,frame,, ]{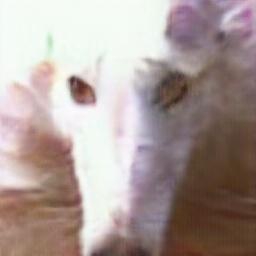} &
        \includegraphics[tSize,frame,, ]{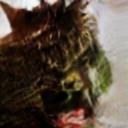} &
        \includegraphics[tSize,frame,, ]{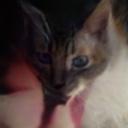}\\

        \includegraphics[tSize,frame,, ]{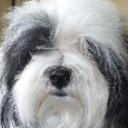} &
        \includegraphics[tSize,frame,, ]{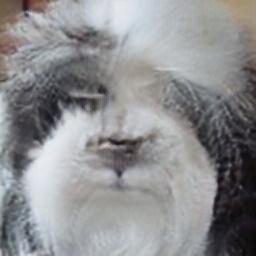} &
        \includegraphics[tSize,frame,, ]{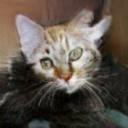} &
        \includegraphics[tSize,frame,, ]{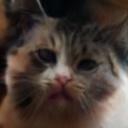}\\

        \includegraphics[tSize,frame,, ]{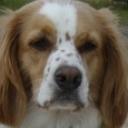} &
        \includegraphics[tSize,frame,, ]{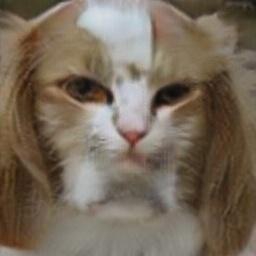} &
        \includegraphics[tSize,frame,, ]{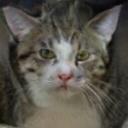} &
        \includegraphics[tSize,frame,, ]{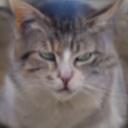}\\
        \tableTitleSize{Input} & \tableTitleSize{CycleGAN} & \tableTitleSize{DiscoGAN} & \tableTitleSize{Ours} \\

\end{tabular}%
\end{table}

\paragraph{Pets in the Wild.}
To demonstrate our network on unaligned data, we evaluate on the Kaggle cat and dog dataset from Microsoft Research~\cite{elson2007catdog}. This dataset contains 12,500 images of each species. The intended purpose of the dataset is to classify cat images from dog images, and so it contains many animal breeds at varying scales, lighting conditions, poses, backgrounds, and occlusion factors.

When translating between cats and dogs (Table \ref{fig:resultsPetsInWild}), the network is able to change both the local features such as the addition and removal of fur and whiskers, plus the larger shape deformation required to fool the discriminator, such as growing a snout. 
Most errors in this domain come from the generator failing to identify an animal from the background, such as forgetting the rear or tail of the animal. 
Sometimes the generator may fail to identify the animal at all. 

We also translate between humans and cats. Table \ref{fig:resultsCatsToHumans} demonstrates how our architecture handles large scale translation with these two variable data distributions. 
Our failure cases are approximately the same as that of the cats to dogs translation, with some promising results. Overall, we translate a surprising degree of shape deformation even when we might not expect this to be possible.

\paragraph{Supplemental Datasets.}
We also tested our approach on existing datasets used in the CycleGAN paper (maps to satellite imagery, horses to zebras, and apples to oranges).
These mappings focus on appearance transfer and require less shape deformation; please see our supplemental material to verify that our approach can handle this setting as well.

\begin{table}[t]
        
    \makeatletter
    \define@key{Gin}{tSize}[true]{%
        \edef\@tempa{{Gin}{width=.1\linewidth, keepaspectratio}}%
        \expandafter\setkeys\@tempa
    }
    \makeatother
    \caption{Pets in the Wild: Between dogs and cats, our approach is able to generate shape transforms across pose and appearance variation.}
    \label{fig:resultsPetsInWild}
    \setlength{\tabcolsep}{0.025\linewidth}
    \renewcommand{\arraystretch}{0.001}%
    \providecommand\aport{}
    \newcommand{\tableTitleSize}[1]{\tiny{#1}}
    \label{tab:catdog}
    
    \centering
    \begin{tabular}[c]{c c}
    \centering
    \rotatebox[origin=c]{90}{ {\large Cat\textrightarrow Dog}}
    \setlength{\tabcolsep}{0.00em}%
    \begin{tabular}[c]{c c c c}
        \tableTitleSize{Input} & \tableTitleSize{CycleGAN} & \tableTitleSize{DiscoGAN} & \tableTitleSize{Ours} \\

        \includegraphics[tSize,frame,]{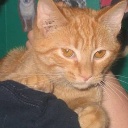} &
        \includegraphics[tSize,frame,]{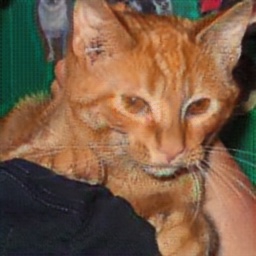} &
        \includegraphics[tSize,frame,, ]{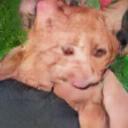} &
        \includegraphics[tSize,frame,]{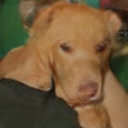}\\

        \includegraphics[tSize,frame,]{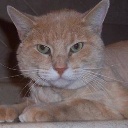}&
        \includegraphics[tSize,frame,]{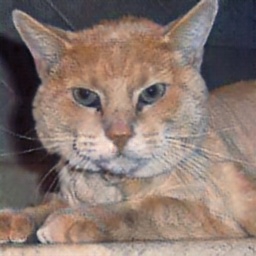}&
        \includegraphics[tSize,frame,, ]{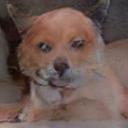}&
        \includegraphics[tSize,frame,]{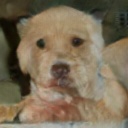} \\

        \includegraphics[tSize,frame,]{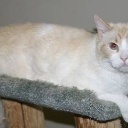}&
        \includegraphics[tSize,frame,]{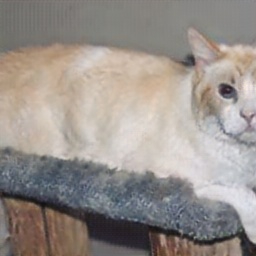}&
        \includegraphics[tSize,frame,, ]{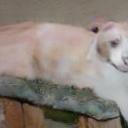}&
        \includegraphics[tSize,frame,]{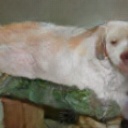} \\

        \includegraphics[tSize,frame,]{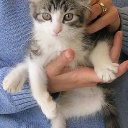}&
        \includegraphics[tSize,frame,]{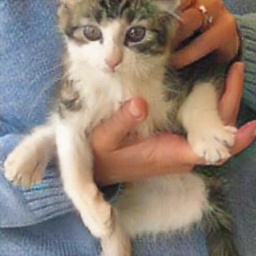}&
        \includegraphics[tSize,frame,, ]{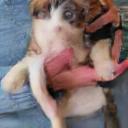}&
        \includegraphics[tSize,frame,]{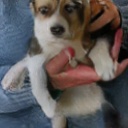} \\

        \includegraphics[tSize,frame,]{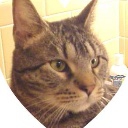}&
        \includegraphics[tSize,frame,]{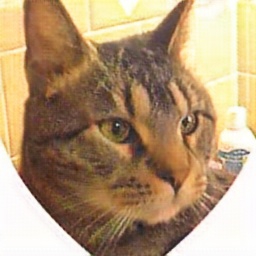}&
        \includegraphics[tSize,frame,, ]{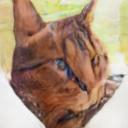}&
        \includegraphics[tSize,frame,]{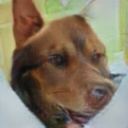} \\

        \includegraphics[tSize,frame,]{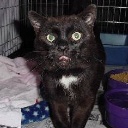}&
        \includegraphics[tSize,frame,]{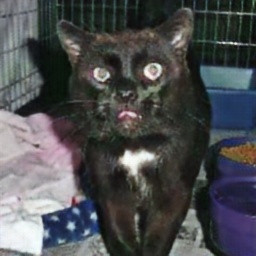}&
        \includegraphics[tSize,frame,, ]{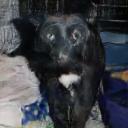}&
        \includegraphics[tSize,frame,]{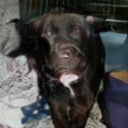} \\

        \includegraphics[tSize,frame,]{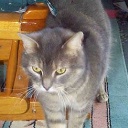}&
        \includegraphics[tSize,frame,]{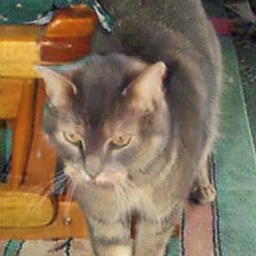}&
        \includegraphics[tSize,frame,, ]{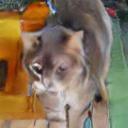}&
        \includegraphics[tSize,frame,]{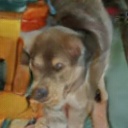} \\

\end{tabular}&
\setlength{\tabcolsep}{0.00em}%
\begin{tabular}[c]{c c c c}
    \tableTitleSize{Input} & \tableTitleSize{CycleGAN} & \tableTitleSize{DiscoGAN} & \tableTitleSize{Ours} \\
    
    \includegraphics[tSize,frame,]{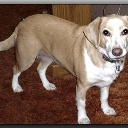} &
    \includegraphics[tSize,frame,]{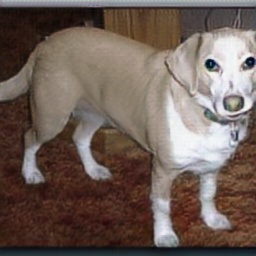} &
    \includegraphics[tSize,frame,, ]{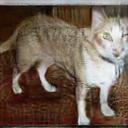} &
    \includegraphics[tSize,frame,]{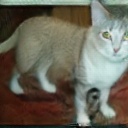}\\

    \includegraphics[tSize,frame,]{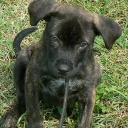}&
    \includegraphics[tSize,frame,]{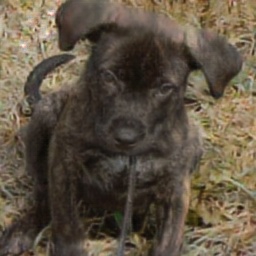}&
    \includegraphics[tSize,frame,, ]{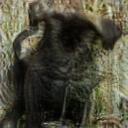} &
    \includegraphics[tSize,frame,]{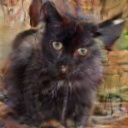} \\

    \includegraphics[tSize,frame,]{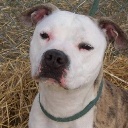}&
    \includegraphics[tSize,frame,]{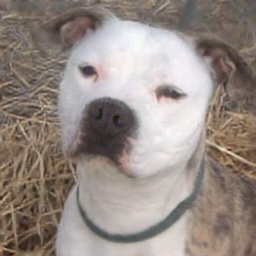}&
    \includegraphics[tSize,frame,, ]{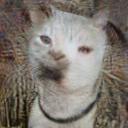} &
    \includegraphics[tSize,frame,]{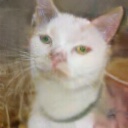} \\

    \includegraphics[tSize,frame,]{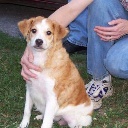}&
    \includegraphics[tSize,frame,]{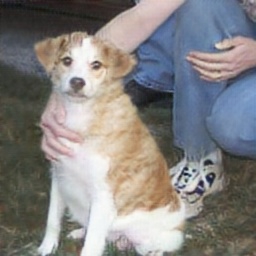}&
    \includegraphics[tSize,frame,, ]{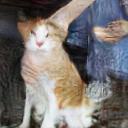} &
    \includegraphics[tSize,frame,]{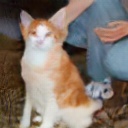} \\

    \includegraphics[tSize,frame,]{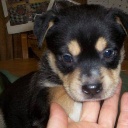}&
    \includegraphics[tSize,frame,]{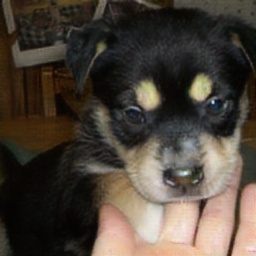}&
    \includegraphics[tSize,frame,, ]{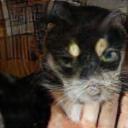} &
    \includegraphics[tSize,frame,]{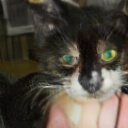} \\

    \includegraphics[tSize,frame,]{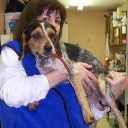}&
    \includegraphics[tSize,frame,]{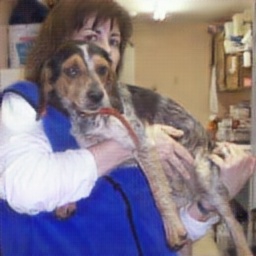}&
    \includegraphics[tSize,frame,, ]{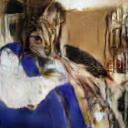} &
    \includegraphics[tSize,frame,]{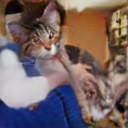} \\

    \includegraphics[tSize,frame,]{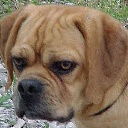}&
    \includegraphics[tSize,frame,]{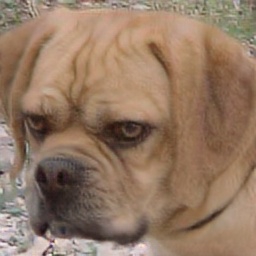}&
    \includegraphics[tSize,frame,, ]{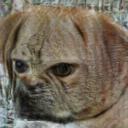} &
    \includegraphics[tSize,frame,]{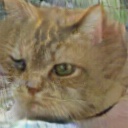}\\
\end{tabular}
\rotatebox[origin=c]{-90}{ {\large Dog\textrightarrow Cat }}
\end{tabular}%
\end{table}

\begin{table}[H]
        
    \makeatletter
    \define@key{Gin}{tSize}[true]{%
        \edef\@tempa{{Gin}{width=0.1175\linewidth, keepaspectratio}}%
        \expandafter\setkeys\@tempa
    }
    \makeatother
    \caption{Human and Pet Faces: As a challenge, we try to map cats to humans and humans to cats. Pose is reliably translated; semantic appearance such as hair color is sometimes translated; but some inputs still fail (bottom left).}
    \setlength{\tabcolsep}{0.05pt}
    \renewcommand{\arraystretch}{0.05}%
    \providecommand\aport{}
    \newcommand{\tableTitleSize}[1]{\footnotesize{#1}}
    
    \centering
    \begin{tabular}[c]{c c c c}
        \tableTitleSize{Input} & \tableTitleSize{Output}  \\
        
        \includegraphics[tSize,frame]{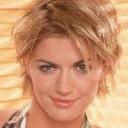} &
        \includegraphics[tSize,frame]{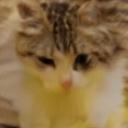} \\
        
        \includegraphics[tSize,frame]{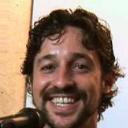} &
        \includegraphics[tSize,frame]{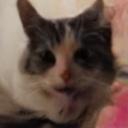} \\
        
        \includegraphics[tSize,frame]{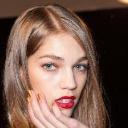} &
        \includegraphics[tSize,frame]{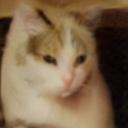} \\
        
        \includegraphics[tSize,frame]{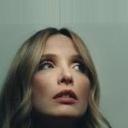} &
        \includegraphics[tSize,frame]{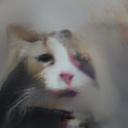} \\
        
        \includegraphics[tSize,frame]{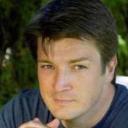} &
        \includegraphics[tSize,frame]{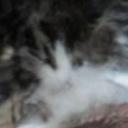} \\

    \end{tabular}%
    \begin{tabular}[c]{c c c c}
        \tableTitleSize{Input} & \tableTitleSize{Output}  \\

        \includegraphics[tSize,frame]{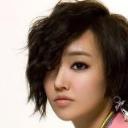} &
        \includegraphics[tSize,frame]{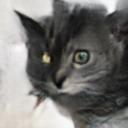} \\      
        
        \includegraphics[tSize,frame]{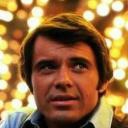} &
        \includegraphics[tSize,frame]{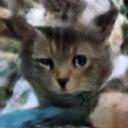} \\
        
        \includegraphics[tSize,frame]{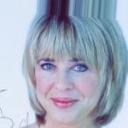} &
        \includegraphics[tSize,frame]{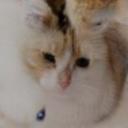} \\
        
        \includegraphics[tSize,frame]{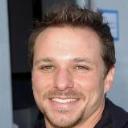} &
        \includegraphics[tSize,frame]{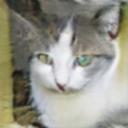} \\
        
        \includegraphics[tSize,frame]{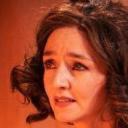} &
        \includegraphics[tSize,frame]{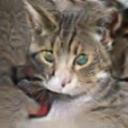} \\
    \end{tabular}%
       \begin{tabular}[c]{c c c c}
        \tableTitleSize{Input} & \tableTitleSize{Output}  \\

        \includegraphics[tSize,frame]{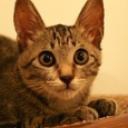} &
        \includegraphics[tSize,frame]{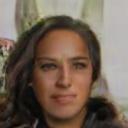} \\
        
        \includegraphics[tSize,frame]{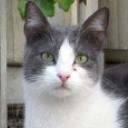} &
        \includegraphics[tSize,frame]{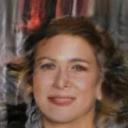} \\

        \includegraphics[tSize,frame]{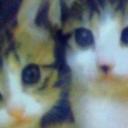} &
        \includegraphics[tSize,frame]{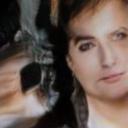} \\
        
        \includegraphics[tSize,frame]{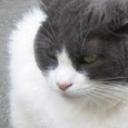} &
        \includegraphics[tSize,frame]{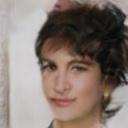} \\
        
        \includegraphics[tSize,frame]{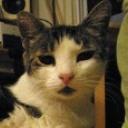} &
        \includegraphics[tSize,frame]{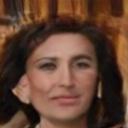} 

    \end{tabular}%
    \begin{tabular}[c]{c c c c}
        \tableTitleSize{Input} & \tableTitleSize{Output}  \\

        \includegraphics[tSize,frame]{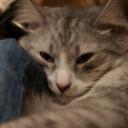} &
        \includegraphics[tSize,frame]{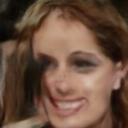} \\
        
        \includegraphics[tSize,frame]{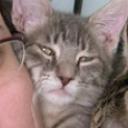} &
        \includegraphics[tSize,frame]{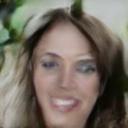} \\      
        
        \includegraphics[tSize,frame]{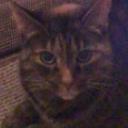} &
        \includegraphics[tSize,frame]{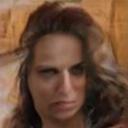} \\
        
        \includegraphics[tSize,frame]{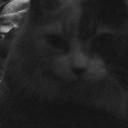} &
        \includegraphics[tSize,frame]{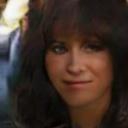} \\
        
        \includegraphics[tSize,frame]{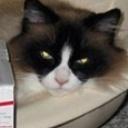} &
        \includegraphics[tSize,frame]{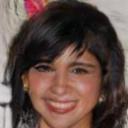} \\
    \end{tabular}%
    \label{fig:resultsCatsToHumans}
\end{table}

\begin{table}[t]
\centering
\caption{Percentage of pixels classified in translated images via CycleGAN, DiscoGAN, and our algorithm (with design choices). Target classes are in \cstress{blue}.}
\label{tab:AblationCatDogBody}
\vspace{-1.5mm}
\begin{tabular}{l rrrr rrrr}
\toprule
Class (\%)  & \multicolumn{4}{c}{Cat\textrightarrow Dog} &  \multicolumn{4}{c}{Dog\textrightarrow Cat } \\
\cmidrule(lr){2-5} \cmidrule(lr){6-9} 
Networks & Cat & \cstress{\textbf{Dog}} & Person & Other & \cstress{\textbf{Cat}} & Dog & Person & Other \\

\midrule
Initial Domain & 100.00 & \cstress{0.00} & 0.00 & 0.00 & \cstress{0.00} & 98.49 & 1.51 & 0.00 \\

\midrule
CycleGAN & 99.99 & \cstress{0.01} & 0.00 & 0.00 & \cstress{2.67} & 97.27 & 0.06 & 0.00 \\
DiscoGAN & 24.37 & \cstress{75.38} & 0.25 & 0.00 & \cstress{96.95} & 0.00 & 2.71 & 0.34 \\
Ours w/ L1 & 100.00 & \cstress{0.00} & 0.00 & 0.00 & \cstress{0.00} & 0.00 & 0.00 & 100.00 \\
Ours w/o feature match loss & 5.03 & \cstress{93.64} & 0.81 & 0.53 & \cstress{85.62} & 14.15 & 0.00 & 0.23 \\

Ours w/ fully conn.~discrim. & 6.11 & \cstress{93.60} & 0.29 & 0.00 & \cstress{91.41} & 8.45 & 0.03 & 0.10 \\
Ours w/ patch discrim. & 46.02 & \cstress{42.90} & 0.05 & 11.03 & \cstress{91.77} & 8.22 & 0.00 & 0.01 \\
Ours (dilated discrim.) & 1.00 & \cstress{\textbf{98.57}} & 0.41 & 0.02 & \cstress{\textbf{100.00}} & 0.00 & 0.00 & 0.00 \\
\bottomrule
\end{tabular}
\vspace{-2mm}
\end{table}

\begin{table}[t]
    \centering
    \setlength{\twidth}{0.135\linewidth}
    \begin{tabular}{c|cc|cc|cc}
        Input & DiscoGAN & & CycleGAN & & Ours & \\
        \includegraphics[width=\twidth,frame] {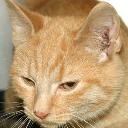} & 
        \includegraphics[width=\twidth,frame] {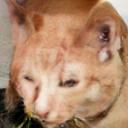} &
        \includegraphics[width=\twidth,frame]{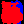} &
        \includegraphics[width=\twidth,frame]{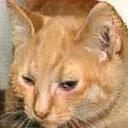} &
        \includegraphics[width=\twidth,frame]{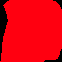} &
        \includegraphics[width=\twidth,frame]{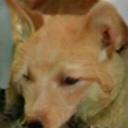} &
        \includegraphics[width=\twidth,frame]{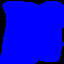}

    \end{tabular}
    \caption{Example segmentation masks from DeepLabV3 for Table~\ref{tab:AblationCatDogBody} for Cat\textrightarrow Dog. {\color{red}{Red}} denotes the cat class, and {\color{blue}{blue}} denotes the intended dog class.}
    \label{tab:segDemo}
    \vspace{-0.5cm}
\end{table}

\begin{table}[t]
\centering
\setlength{\twidth}{0.15\linewidth}

\newcommand{\tableTitleSize}[1]{\scriptsize{#1}}
\newcolumntype{P}[1]{>{\centering\arraybackslash}p{#1}}

\caption{In qualitative comparisons, GANimorph outperforms all of its ablated versions. For instance, our approach better resolves fine details (e.g., second row, cat eyes) while also better translating the overall shape (e.g., last row, cat nose and ears).}
\label{tab:AblationCatDogBodyQualitative}
\vspace{-2mm}
\setlength\tabcolsep{0.5pt}
\renewcommand{\arraystretch}{0.05}

\begin{tabular}{*{6}{P{\twidth}}}
    \tableTitleSize{Input} & \tableTitleSize{No FM Loss} & \tableTitleSize{L1 Loss} & \tableTitleSize{Patch \mbox{Discrim}} & \tableTitleSize{FC \mbox{Discrim}} & \tableTitleSize{Ours}
    \\
    \includegraphics[frame,width=\twidth,]{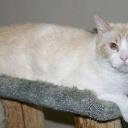} &
    \includegraphics[frame,width=\twidth,]{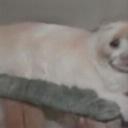} &
    \includegraphics[frame,width=\twidth,]{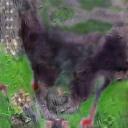} &
    \includegraphics[frame,width=\twidth,]{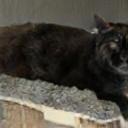} &
    \includegraphics[frame,width=\twidth,]{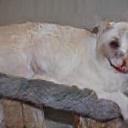} &
    \includegraphics[frame,width=\twidth,]{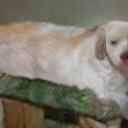}
    \\
    \includegraphics[frame,width=\twidth,]{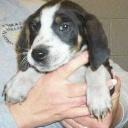} &
    \includegraphics[frame,width=\twidth,]{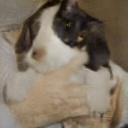} &
    \includegraphics[frame,width=\twidth,]{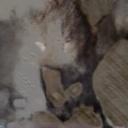} &
    \includegraphics[frame,width=\twidth,]{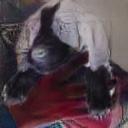} &
    \includegraphics[frame,width=\twidth,]{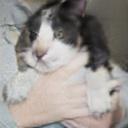} &
    \includegraphics[frame,width=\twidth,]{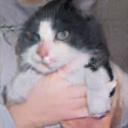}
    \\
    \includegraphics[frame,width=\twidth,]{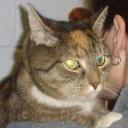} &
    \includegraphics[frame,width=\twidth,]{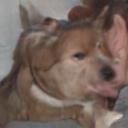} &
    \includegraphics[frame,width=\twidth,]{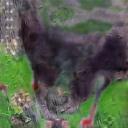} &
    \includegraphics[frame,width=\twidth,]{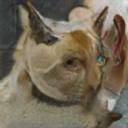} &
    \includegraphics[frame,width=\twidth,]{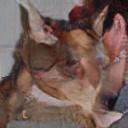} &
    \includegraphics[frame,width=\twidth,]{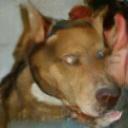}
    \\
    \includegraphics[frame,width=\twidth,]{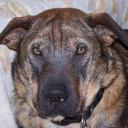} &
    \includegraphics[frame,width=\twidth,]{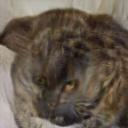} &
    \includegraphics[frame,width=\twidth,]{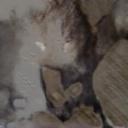} &
    \includegraphics[frame,width=\twidth,]{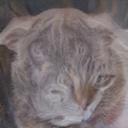} &
    \includegraphics[frame,width=\twidth,]{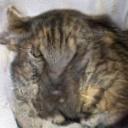} &
    \includegraphics[frame,width=\twidth,]{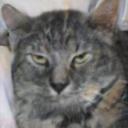}
\end{tabular}

\vspace{-2mm}
\end{table}

\subsection{Quantitative Study}

To quantify GANimorph's translation ability, we consider classification-based metrics to detect class change, e.g., whether a cat was successfully translated into a dog. Since there is no per pixel ground truth in this task for any real-world datasets, we cannot use Fully Convolution Score. Using Inception Score~\cite{salimans2016fm} is uninformative since simply outputting the original image would score highly. 

Further, similar to adversarial examples, CycleGAN is able to convince many classification networks that the image is translated even though to a human the image appears untranslated: all CycleGAN results from Table \ref{fig:petfaces} convince both ResNet50~\cite{he2015resnet} and the traditional segmentation network of Zheng et al.~\cite{zheng2015conditional}, even though the image is unsuccessfully translated.

However, semantic segmentation networks that model multi-scale properties can distinguish CycleGAN's `adversarial examples' from true translations, such as DeepLabV3~\cite{chen2017deeplabv3} (trained on PascalVOC 2012 and using dilated convolutions itself). As such, we run each test image through the DeepLabV3 network to generate a segmentation mask.  Then, we compute the percent of non-background-labeled pixels per class, and average across the test set (Table \ref{tab:AblationCatDogBody}). Our approach is able to more fully translate the image in the eyes of the classification network, with images also appearing translated to a human (Table \ref{tab:segDemo}).

\subsection{Ablation Study} We use these quantiative settings for an ablation study (Table \ref{tab:AblationCatDogBody}). 
First, we removed MS-SSIM to leave only L1 ($\mathcal{L}_{\textnormal{SS}}$, Eq.~\ref{eqn:GGANtot}), which causes our network to mode collapse. 
Next, we removed feature match loss, but this decreases both our segmentation consistency and the stability of the network. 
Then, we replaced our dilated discriminator with a patch discriminator. 
However, the patch discriminator cannot use global context, and so the network confuses facial layouts. 
Finally, we replace our dilated discriminator with a fully connected discriminator. 
We see that our generator architecture and loss function allow our network to outperform DiscoGAN even with the same type of discriminator (fully connected). 

Qualitative ablation study results are shown in Table~\ref{tab:AblationCatDogBodyQualitative}. The patch based discriminator translates texture well, but fails to create globally-coherent images. Decreasing the information flow by using a fully-connected discriminator or removing feature match leads to better results. Maximizing the information flow ultimately leads to the best results (last column). Using L1 instead of a perceptual cyclic loss term leads to mode collapse.

\section{Discussion}

There exists a trade off in the relative weighting of the cyclic loss. A higher cyclic loss term weight $\lambda_{cyc}$ will prevent significant shape change and weaken the generator's ability to adapt to the discriminator. Setting it too low will cause the collapse of the network and prevent any meaningful mapping from existing between domains. For instance, the network can easily hallucinate objects in the other domain if the reconstruction loss is too low. Likewise, setting it too high will prevent the network from deforming the shape properly. As such, an architecture which could modify the weight of this term at test time would prove valuable for user control over how much deformation to allow. 

One counter-intuitive result we discovered is that in domains with little variety, the mappings can lose semantic meaning (see supplemental material). One example of a failed mapping was from celebA to bitmoji faces \cite{taigman2016emoji}. Many attributes were lost, including pose, and the mapping fell back to pseudo-stegano\-graphic encoding of the faces~\cite{chu2017cyclesteg}. For example, background information would be encoded in color gradients of hair styles, and minor variations in the width of the eyes were used similarly. As such, the cyclic loss limits the ability of the network to abstract relevant details. Approaches such as relying on mapping the variance within each dataset, similar to Benaim et al.~\cite{benaim2017distancegan}, may prove an effective means of ensuring the variance in either domain is maintained. We found that this term over-constrained the amount of shape change in the target domain; however, this may be worth further investigation.

Finally, trying to learn each domain simultaneously may also prove an effective way to increase the accuracy of image translation. 
Doing so allows the discriminator(s) and generator to learn how to better determine and transform regions of interest for either network. 
Better results might be obtained by mapping between multiple domains using parameter-efficient networks (e.g., StarGAN~\cite{choi2017stargan}).

\vspace{-0.25cm}
\section{Conclusion}
\vspace{-0.1cm}
We have demonstrated that reframing the discriminator's role as a semantic segmenter allows greater shape change with less image artifacts.
Further, that training with a perceptual cyclic loss and that adding explicit multi-scale features both help the network to translate more complex shape deformation.
Finally, that training techniques such as feature matching loss and scheduled loss normalization can increase the performance of translation networks.
In summary, our architecture and training changes allow the network to go beyond simple texture transfer and improve shape deformation. 
This lets our GANimorph system perform challenging translations such as from human to anime and feline faces, and from cats to dogs. 
The source code to our GANimorph system and all datasets are online: \href{https://github.com/brownvc/ganimorph/}{https://github.com/brownvc/ganimorph/}.

\vspace{-0.1cm}
\paragraph{Acknowledgement:} Kwang~In~Kim thanks RCUK EP/M023281/1, and Aaron Gokaslan and James Tompkin thank NVIDIA Corporation.

\bibliographystyle{splncs04}
\bibliography{bibliography}

\clearpage
\appendix
\section{Appendix}

\subsection{Optimization and Loss Parameters}
For convenience, we list all optimization and loss parameters in Table \ref{tab:params}.

\begin{table}[b]
    \centering
    \caption{\emph{Left:} Optimization parameter values. \emph{Right:} Loss hyperparameter values.}
    \label{tab:params}
    \begin{tabular}{l r}
        \toprule
        Optimization term & Value \\
        \midrule
        Learning rate & 2e-4  \\
        Minibatch size & 16 \\
        Residual Blocks & 3 \\
        Residual Merge Op. & Concat \\
        \midrule
        Optimizer & ADAM~\cite{Kingma2014ADAM} \\
        Momentum $\beta_1$ & 0.95 \\
        $\beta_2$ (ADAM) & 0.999 \\
        \midrule
        $\beta$ ($\mathcal{L}_{\textnormal{moavg}}$) & 0.99 \\
        $\eta$ & $10^{-10}$ \\
        $s$ & 200 \\
        \bottomrule
    \end{tabular}
    \quad
    \begin{tabular}{l r}
        \toprule
        Hyperparameter \hspace{0.25cm} & Value\\
        \midrule
        $\lambda_{\textnormal{GAN}}$ & 0.49 \\
        $\lambda_{\textnormal{FM}}$  & 0.21 \\
        $\lambda_{\textnormal{CYC}}$ & 0.30 \\
        $\lambda_{\textnormal{SS}}$  & 0.70 \\
        $\lambda_{\textnormal{L1}}$  & 0.30 \\
        \bottomrule
    \end{tabular}
\end{table}

\begin{table}[t]
\centering
\caption{Number of iterations per dataset, with how often the discriminator was updated in interations.}
\label{tab:updatesteps}
\begin{tabular}{l r r}
    \toprule\
    Dataset & Iterations & \hspace{0.15cm} Discrim.~every \\
    \midrule           
    Anime (Danbooru)    & 200,000  & 2 \\
    Anime (Getchu)      & 200,000  & 1 \\
    Doll                & 100,000  & 2 \\
    Cat/dog faces       & 150,000  & 2 \\
    Cat/dog bodies      & 300,000  & 2 \\
    Toy dataset         & 150,000  & 1 \\
    CycleGAN datasets   & 200,000  & 1 \\
    \bottomrule
\end{tabular}
\end{table}

\subsection{Network}

We use 64 filters for the first layer of the generator and 128 filters for the first layer of the generator. Then, for subsequent layers, we double the number of filters for every downsampling stride of two (main paper, Figure 2). The stride is two for all downsampling layers. We do not increase the number of filters for dilated convolutions. Likewise, we decrease the number of filter for each transposed convolution by a factor of two. We also linearly decay the learning rate from 150k steps onwards, to approach 0 at 300k steps. Table \ref{tab:updatesteps} lists the number of update steps computed per dataset.

\subsection{Existing Dataset Comparison}

We compute comparisons with our method on existing unsupervised image-to-image translation datasets. We trained CycleGAN \cite{zhuICCV2017} and our architecture on the same datasets for the same number of iterations.

\textit{Satellite to Google Maps.} In Table \ref{tab:map2sat}, we compare results for translating satellite imagery to Google Maps and vice versa. The Google Maps dataset carries less information overall than the satellite dataset, so this task requires the network to `encode detail in plain sight' when translating in the Google Maps to satellite direction \cite{chu2017cyclesteg}. Generally, our network produces comparable results, with some differences in ambiguous cases. This provides evidence that our network is also able to solve tasks with very little shape deformation.

\textit{Apples to Oranges.} Table \ref{tab:apple2orange} shows the results. With less shape change between elements in the domain, our approach produces comparable results to CycleGAN.

\textit{Horse to Zebra.} Table \ref{tab:horse2zebra} shows the results, and vice versa. In general, the results between the two techniques are comparable. One improvement that our method is able to make is to better maintain global orientation of stripes, e.g., in column 1, rows 2 and 3, we see that our method places horizontal stripes on the rear of the zebra, which rotate over the body of the animal to vertical stripes on the neck. Overall, this is still a hard problem, and many examples show artifacts for both techniques.

\textit{Getchu to CelebA.} First, we collected dataset from Getchu.com~\cite{jin2017makegirlsmoe}, which consists of professionally-drawn visual novel characters from 1995--2017. Table \ref{tab:getchu} shows the results, including failure cases. When translating from anime to CelebA, CycleGAN often has trouble to substantially change the shape of the character's face, often simply attempting to replace anime shading with skin tones (see row 1 column 1). Our network is better able to map both pose and facial structure, while also blending skin tone more appropriately. One limitation is that the Getchu dataset has different attribute variance, e.g., more pink and purple hair, and less diversity in skin colors. This restricts the ability of the methods to transfer attributes successfully.

\textit{Limitation: Human to Bitmoji.} In Table \ref{tab:bitmoji}, we show a significant failure cases of our network in the Human to Bitmoji task. Our network mode collapses and often fails to properly match the pose of the target distribution. This is interesting because our reconstructions for these samples are almost identical to the input image, implying that our network is able to encode the entire database into this single sample using almost imperceptible differences in the image \cite{chu2017cyclesteg}. The failure is likely caused by the simplicity of the bitmoji domain, e.g., uniform skin color. The anime and other domains contain enough for information to prevent the network from learning such a steganographic encoding.

\begin{table}[tbp]  
    \makeatletter
    \define@key{Gin}{tSize}[true]{%
        \edef\@tempa{{Gin}{width=.11\linewidth, keepaspectratio}}%
        \expandafter\setkeys\@tempa
    }
    \makeatother
    \caption{Satellite to Google Maps. Generally, our network produces comparable results, with some differences in ambiguous cases. This provides evidence that our network is also able to solve tasks with very little shape deformation.}
    \setlength{\tabcolsep}{0.00em}
    \renewcommand{\arraystretch}{0.01}%
    \providecommand\aport{}
    \newcommand{\tableTitleSize}[1]{\tiny{#1}}
    
    \centering
\begin{tabular}{c c c}
        \tableTitleSize{Input} & \tableTitleSize{CycleGAN} & \tableTitleSize{Ours} \\

        \includegraphics[tSize,frame,]{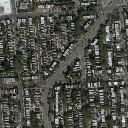} &
        \includegraphics[tSize,frame,]{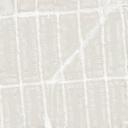} &
        \includegraphics[tSize,frame,]{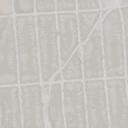}\\
        
        \includegraphics[tSize,frame,]{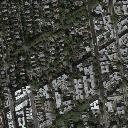} &
        \includegraphics[tSize,frame,]{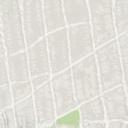} &
        \includegraphics[tSize,frame,]{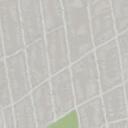}\\    
        
        \includegraphics[tSize,frame,]{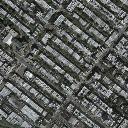} &
        \includegraphics[tSize,frame,]{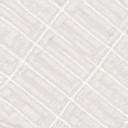} &
        \includegraphics[tSize,frame,]{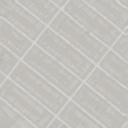}\\
        
        \tableTitleSize{Input} & \tableTitleSize{CycleGAN} & \tableTitleSize{Ours} \\
 \end{tabular}%
 \begin{tabular}{c c c}
        \tableTitleSize{Input} & \tableTitleSize{CycleGAN} & \tableTitleSize{Ours} \\

        \includegraphics[tSize,frame,]{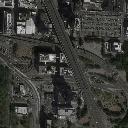} &
        \includegraphics[tSize,frame,]{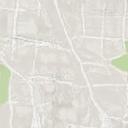} &
        \includegraphics[tSize,frame,]{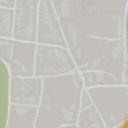}\\
        
        \includegraphics[tSize,frame,]{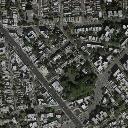} &
        \includegraphics[tSize,frame,]{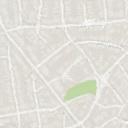} &
        \includegraphics[tSize,frame,]{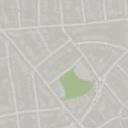}\\    
        
        \includegraphics[tSize,frame,]{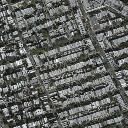} &
        \includegraphics[tSize,frame,]{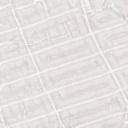} &
        \includegraphics[tSize,frame,]{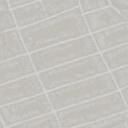}\\
        
        \tableTitleSize{Input} & \tableTitleSize{CycleGAN} & \tableTitleSize{Ours} \\
 \end{tabular}%
 \begin{tabular}{c c c}
        \tableTitleSize{Input} & \tableTitleSize{CycleGAN} & \tableTitleSize{Ours} \\

        \includegraphics[tSize,frame,]{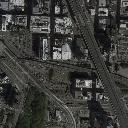} &
        \includegraphics[tSize,frame,]{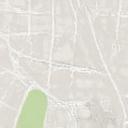} &
        \includegraphics[tSize,frame,]{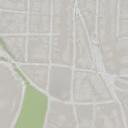}\\
        
        \includegraphics[tSize,frame,]{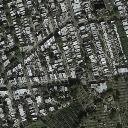} &
        \includegraphics[tSize,frame,]{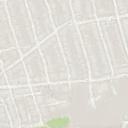} &
        \includegraphics[tSize,frame,]{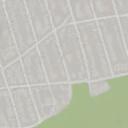}\\    
        
        \includegraphics[tSize,frame,]{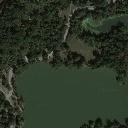} &
        \includegraphics[tSize,frame,]{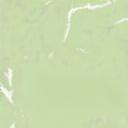} &
        \includegraphics[tSize,frame,]{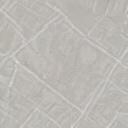}\\
        
        \tableTitleSize{Input} & \tableTitleSize{CycleGAN} & \tableTitleSize{Ours} \\
\end{tabular}%

\begin{tabular}{c c c}

        \includegraphics[tSize,frame,]{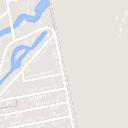} &
        \includegraphics[tSize,frame,]{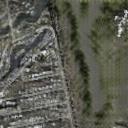} &
        \includegraphics[tSize,frame,]{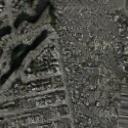}\\
        
        \includegraphics[tSize,frame,]{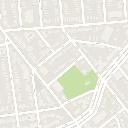} &
        \includegraphics[tSize,frame,]{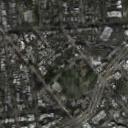} &
        \includegraphics[tSize,frame,]{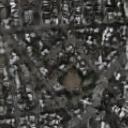}\\    
        
        \includegraphics[tSize,frame,]{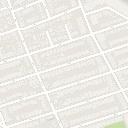} &
        \includegraphics[tSize,frame,]{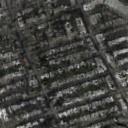} &
        \includegraphics[tSize,frame,]{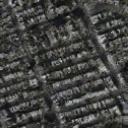}\\
        
        \tableTitleSize{Input} & \tableTitleSize{CycleGAN} & \tableTitleSize{Ours} \\
 \end{tabular}%
  \begin{tabular}{c c c}

        \includegraphics[tSize,frame,]{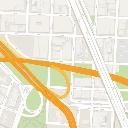} &
        \includegraphics[tSize,frame,]{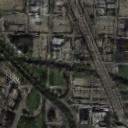} &
        \includegraphics[tSize,frame,]{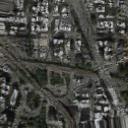}\\
        
        \includegraphics[tSize,frame,]{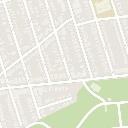} &
        \includegraphics[tSize,frame,]{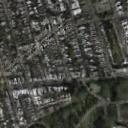} &
        \includegraphics[tSize,frame,]{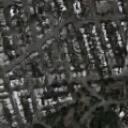}\\    
        
        \includegraphics[tSize,frame,]{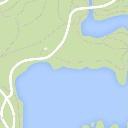} &
        \includegraphics[tSize,frame,]{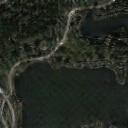} &
        \includegraphics[tSize,frame,]{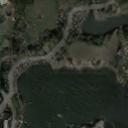}\\
        
        \tableTitleSize{Input} & \tableTitleSize{CycleGAN} & \tableTitleSize{Ours} \\
 \end{tabular}%
 \begin{tabular}{c c c}

        \includegraphics[tSize,frame,]{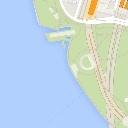} &
        \includegraphics[tSize,frame,]{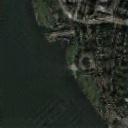} &
        \includegraphics[tSize,frame,]{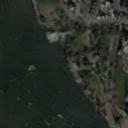}\\
        
        \includegraphics[tSize,frame,]{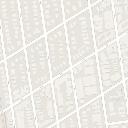} &
        \includegraphics[tSize,frame,]{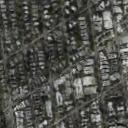} &
        \includegraphics[tSize,frame,]{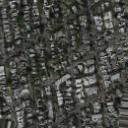}\\
        
        \includegraphics[tSize,frame,]{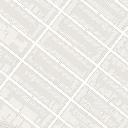} &
        \includegraphics[tSize,frame,]{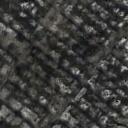} &
        \includegraphics[tSize,frame,]{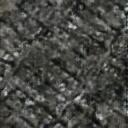}\\
        
        \tableTitleSize{Input} & \tableTitleSize{CycleGAN} & \tableTitleSize{Ours} \\
 \end{tabular}%
 \label{tab:map2sat}
\end{table}

\begin{table}    
    \makeatletter
    \define@key{Gin}{tSize}[true]{%
        \edef\@tempa{{Gin}{width=.11\linewidth, keepaspectratio}}%
        \expandafter\setkeys\@tempa
    }
    \makeatother
    \caption{On Apples to Oranges, results between the two techniques are approximately comparable, though the shapes of the two fruits are already similar.}
    \setlength{\tabcolsep}{0.00em}
    \renewcommand{\arraystretch}{0.01}%
    \providecommand\aport{}
    \newcommand{\tableTitleSize}[1]{\tiny{#1}}
    
    \centering
    \begin{tabular}{c c c}
        \tableTitleSize{Input} & \tableTitleSize{CycleGAN} & \tableTitleSize{Ours} \\

        \includegraphics[tSize,frame,]{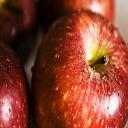} &
        \includegraphics[tSize,frame,, ]{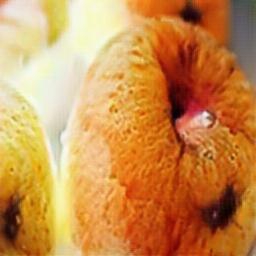} &
        \includegraphics[tSize,frame,]{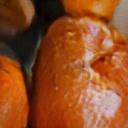}\\
        
        \includegraphics[tSize,frame,]{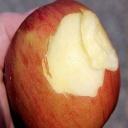} &
        \includegraphics[tSize,frame,, ]{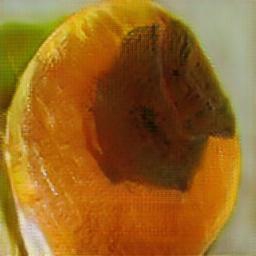} &
        \includegraphics[tSize,frame,]{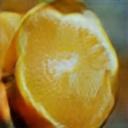}\\
        
        \includegraphics[tSize,frame,]{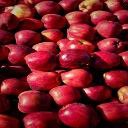} &
        \includegraphics[tSize,frame,, ]{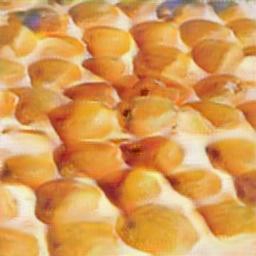} &
        \includegraphics[tSize,frame,]{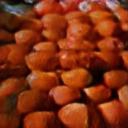}\\
        
        \tableTitleSize{Input} & \tableTitleSize{CycleGAN} & \tableTitleSize{Ours} \\
 \end{tabular}%
 \begin{tabular}{c c c}
        \tableTitleSize{Input} & \tableTitleSize{CycleGAN} & \tableTitleSize{Ours} \\

        \includegraphics[tSize,frame,]{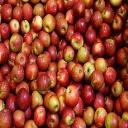} &
        \includegraphics[tSize,frame,, ]{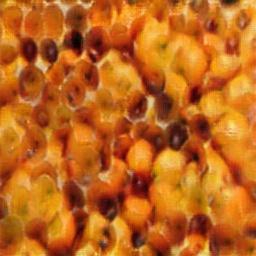} &
        \includegraphics[tSize,frame,]{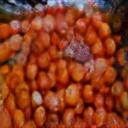}\\
        
        \includegraphics[tSize,frame,]{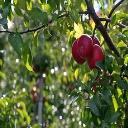} &
        \includegraphics[tSize,frame,, ]{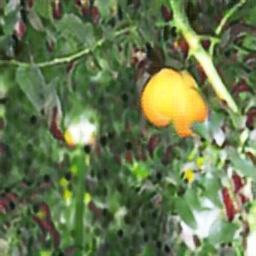} &
        \includegraphics[tSize,frame,]{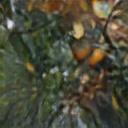}\\
        
        \includegraphics[tSize,frame,]{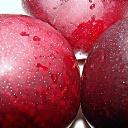} &
        \includegraphics[tSize,frame,, ]{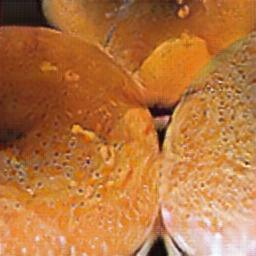} &
        \includegraphics[tSize,frame,]{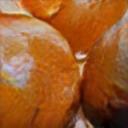}\\
        
        \tableTitleSize{Input} & \tableTitleSize{CycleGAN} & \tableTitleSize{Ours} \\
 \end{tabular}%
  \begin{tabular}{c c c}
        \tableTitleSize{Input} & \tableTitleSize{CycleGAN} & \tableTitleSize{Ours} \\

        \includegraphics[tSize,frame,]{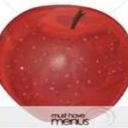} &
        \includegraphics[tSize,frame,, ]{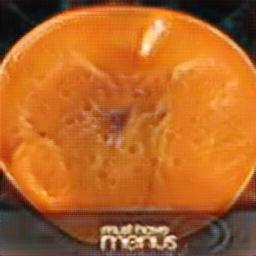} &
        \includegraphics[tSize,frame,]{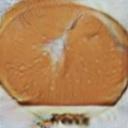}\\
        
        \includegraphics[tSize,frame,]{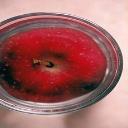} &
        \includegraphics[tSize,frame,, ]{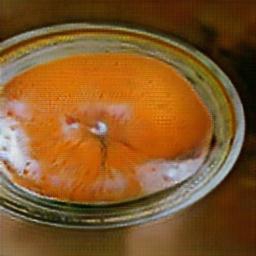} &
        \includegraphics[tSize,frame,]{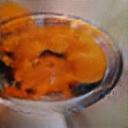}\\
        
        \includegraphics[tSize,frame,]{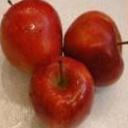} &
        \includegraphics[tSize,frame,, ]{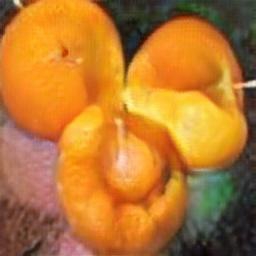} &
        \includegraphics[tSize,frame,]{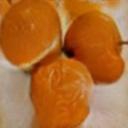}\\

        \tableTitleSize{Input} & \tableTitleSize{CycleGAN} & \tableTitleSize{Ours} \\
 \end{tabular}%
 
 \begin{tabular}{c c c}

        \includegraphics[tSize,frame,]{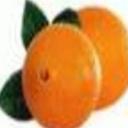} &
        \includegraphics[tSize,frame,, ]{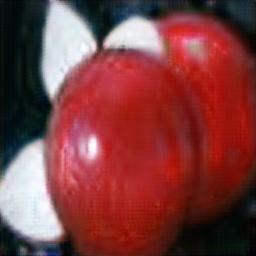} &
        \includegraphics[tSize,frame,]{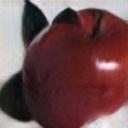}\\
        
        \includegraphics[tSize,frame,]{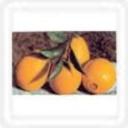} &
        \includegraphics[tSize,frame,, ]{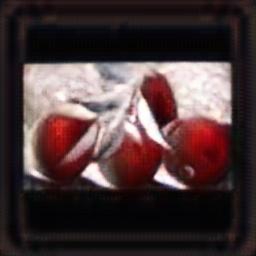} &
        \includegraphics[tSize,frame,]{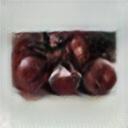}\\
        
        \includegraphics[tSize,frame,]{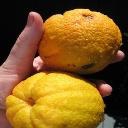} &
        \includegraphics[tSize,frame,, ]{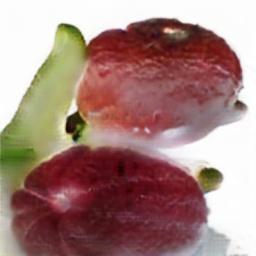} &
        \includegraphics[tSize,frame,]{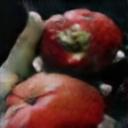}\\
        
        \tableTitleSize{Input} & \tableTitleSize{CycleGAN} & \tableTitleSize{Ours} \\
 \end{tabular}%
  \begin{tabular}{c c c}

        \includegraphics[tSize,frame,]{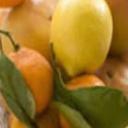} &
        \includegraphics[tSize,frame,, ]{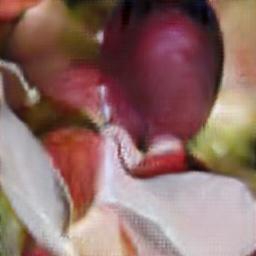} &
        \includegraphics[tSize,frame,]{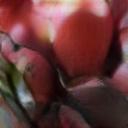}\\
        
        \includegraphics[tSize,frame,]{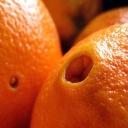} &
        \includegraphics[tSize,frame,, ]{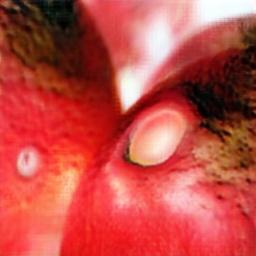} &
        \includegraphics[tSize,frame,]{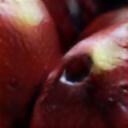}\\
        
        \includegraphics[tSize,frame,]{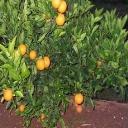} &
        \includegraphics[tSize,frame,, ]{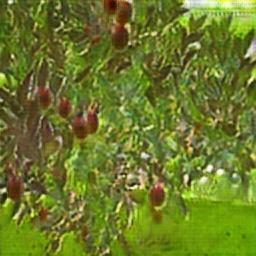} &
        \includegraphics[tSize,frame,]{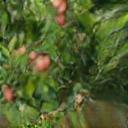}\\
        
        \tableTitleSize{Input} & \tableTitleSize{CycleGAN} & \tableTitleSize{Ours} \\
 \end{tabular}%
   \begin{tabular}{c c c}

        \includegraphics[tSize,frame,]{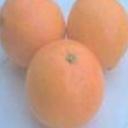} &
        \includegraphics[tSize,frame,, ]{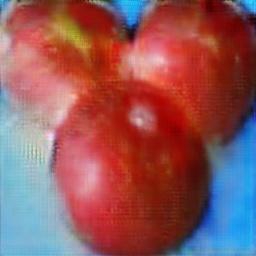} &
        \includegraphics[tSize,frame,]{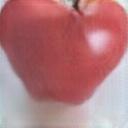}\\
        
        \includegraphics[tSize,frame,]{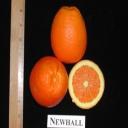} &
        \includegraphics[tSize,frame,, ]{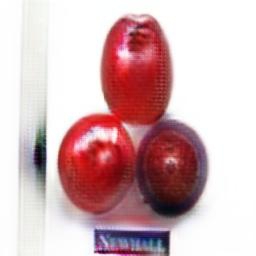} &
        \includegraphics[tSize,frame,]{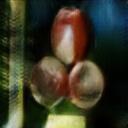}\\
        
        \includegraphics[tSize,frame,]{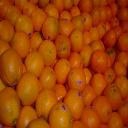} &
        \includegraphics[tSize,frame,, ]{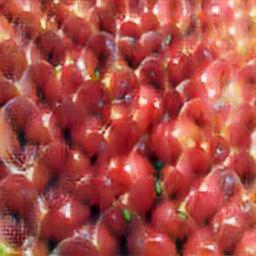} &
        \includegraphics[tSize,frame,]{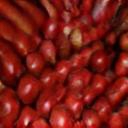}\\
        
        \tableTitleSize{Input} & \tableTitleSize{CycleGAN} & \tableTitleSize{Ours} \\
 \end{tabular}%
 \label{tab:apple2orange}
\end{table}
  
\begin{table}    
    \makeatletter
    \define@key{Gin}{tSize}[true]{%
        \edef\@tempa{{Gin}{width=.14\linewidth, keepaspectratio}}%
        \expandafter\setkeys\@tempa
    }
    \makeatother
    \caption{Our method is able to transfer local zebra texture onto horses comparably to CycleGAN. However, our method is better able to maintain global stripe orientation, e.g., horizontal stripes on the rump, rotating to vertical stripes on the neck. When removing texture to turn a zebra into a horse, our method performs comparably.}
    \setlength{\tabcolsep}{0.00em}
    \renewcommand{\arraystretch}{0.01}%
    \providecommand\aport{}
    \newcommand{\tableTitleSize}[1]{\tiny{#1}}
    
    \centering
\begin{tabular}{c c c}
        \tableTitleSize{Input} & \tableTitleSize{CycleGAN} & \tableTitleSize{Ours} \\
        
        \includegraphics[tSize,frame,]{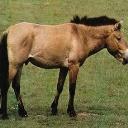} &
        \includegraphics[tSize,frame,]{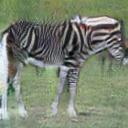} &
        \includegraphics[tSize,frame,]{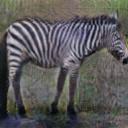}\\
                
        \includegraphics[tSize,frame,]{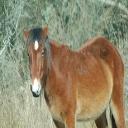} &
        \includegraphics[tSize,frame,]{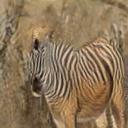} &
        \includegraphics[tSize,frame,]{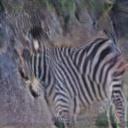}\\
        
        \includegraphics[tSize,frame,]{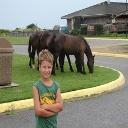} &
        \includegraphics[tSize,frame,]{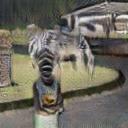} &
        \includegraphics[tSize,frame,]{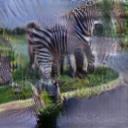}\\
        
        \tableTitleSize{Input} & \tableTitleSize{CycleGAN} & \tableTitleSize{Ours} \\
 \end{tabular}%
 \begin{tabular}{c c c}
        \tableTitleSize{Input} & \tableTitleSize{CycleGAN} & \tableTitleSize{Ours} \\

        \includegraphics[tSize,frame,]{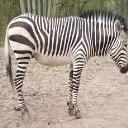} &
        \includegraphics[tSize,frame,]{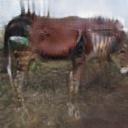} &
        \includegraphics[tSize,frame,]{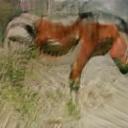}\\
                
        \includegraphics[tSize,frame,]{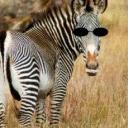} &
        \includegraphics[tSize,frame,]{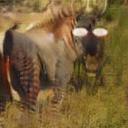} &
        \includegraphics[tSize,frame,]{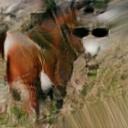}\\
                
        \includegraphics[tSize,frame,]{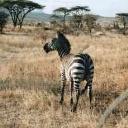} &
        \includegraphics[tSize,frame,]{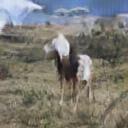} &
        \includegraphics[tSize,frame,]{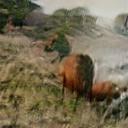}\\
        
        \tableTitleSize{Input} & \tableTitleSize{CycleGAN} & \tableTitleSize{Ours} \\
 \end{tabular}%
 \label{tab:horse2zebra}
\end{table}

\begin{table}
    \makeatletter
    \define@key{Gin}{tSize}[true]{%
        \edef\@tempa{{Gin}{width=.105\linewidth, keepaspectratio}}%
        \expandafter\setkeys\@tempa
    }
    \makeatother
    \caption{Anime to Human (and vice versa) on the Getchu dataset. Our method is more successful at making shape changes, e.g., shrinking the head when translating to anime, or growing it when translating to human. Different attribute variances between the two datasets, e.g., hair or skin color, sometimes prevents attribute transfer.}
    \setlength{\tabcolsep}{0.00em}
    \renewcommand{\arraystretch}{0.01}%
    \providecommand\aport{}
    \newcommand{\tableTitleSize}[1]{\tiny{#1}}
    
    \centering
\begin{tabular}{c c c}
        \tableTitleSize{Input} & \tableTitleSize{CycleGAN} & \tableTitleSize{Ours} \\

        \includegraphics[tSize,frame,]{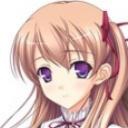} &
        \includegraphics[tSize,frame,]{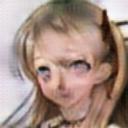} &
        \includegraphics[tSize,frame,]{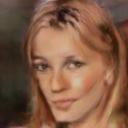}\\
        
        \includegraphics[tSize,frame,]{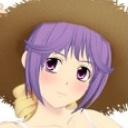} &
        \includegraphics[tSize,frame,]{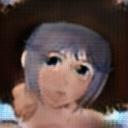} &
        \includegraphics[tSize,frame,]{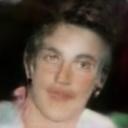}\\
        
        \includegraphics[tSize,frame,]{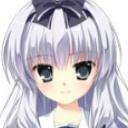} &
        \includegraphics[tSize,frame,]{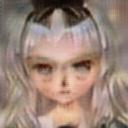} &
        \includegraphics[tSize,frame,]{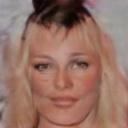}\\
        
        \tableTitleSize{Input} & \tableTitleSize{CycleGAN} & \tableTitleSize{Ours} \\
 \end{tabular}%
\begin{tabular}{c c c}
        \tableTitleSize{Input} & \tableTitleSize{CycleGAN} & \tableTitleSize{Ours} \\

        \includegraphics[tSize,frame,]{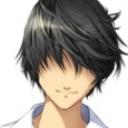} &
        \includegraphics[tSize,frame,]{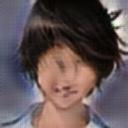} &
        \includegraphics[tSize,frame,]{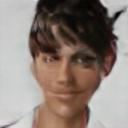}\\
        
        \includegraphics[tSize,frame,]{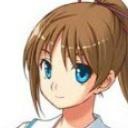} &
        \includegraphics[tSize,frame,]{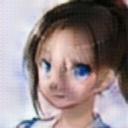} &
        \includegraphics[tSize,frame,]{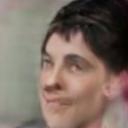}\\
        
        \includegraphics[tSize,frame,]{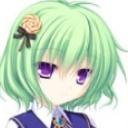} &
        \includegraphics[tSize,frame,]{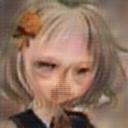} &
        \includegraphics[tSize,frame,]{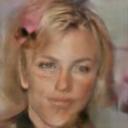}\\
        
        \tableTitleSize{Input} & \tableTitleSize{CycleGAN} & \tableTitleSize{Ours} \\
 \end{tabular}%
 \begin{tabular}{c c c}
        \tableTitleSize{Input} & \tableTitleSize{CycleGAN} & \tableTitleSize{Ours} \\

        \includegraphics[tSize,frame,]{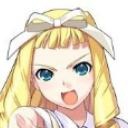} &
        \includegraphics[tSize,frame,]{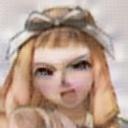} &
        \includegraphics[tSize,frame,]{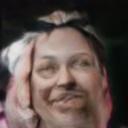}\\
        
        \includegraphics[tSize,frame,]{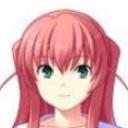} &
        \includegraphics[tSize,frame,]{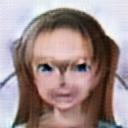} &
        \includegraphics[tSize,frame,]{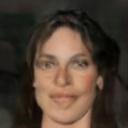}\\
        
        \includegraphics[tSize,frame,]{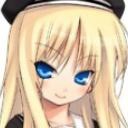} &
        \includegraphics[tSize,frame,]{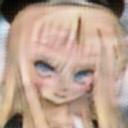} &
        \includegraphics[tSize,frame,]{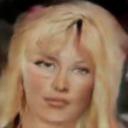}\\
        
        \tableTitleSize{Input} & \tableTitleSize{CycleGAN} & \tableTitleSize{Ours} \\
 \end{tabular}%
 \\
\begin{tabular}{c c c}

        \includegraphics[tSize,frame,]{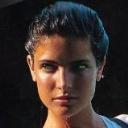} &
        \includegraphics[tSize,frame,]{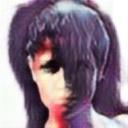} &
        \includegraphics[tSize,frame,]{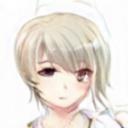}\\
        
        \includegraphics[tSize,frame,]{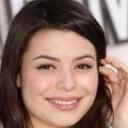} &
        \includegraphics[tSize,frame,]{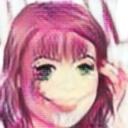} &
        \includegraphics[tSize,frame,]{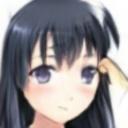}\\
        
        \includegraphics[tSize,frame,]{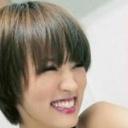} &
        \includegraphics[tSize,frame,]{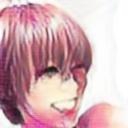} &
        \includegraphics[tSize,frame,]{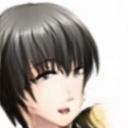}\\
        
        \tableTitleSize{Input} & \tableTitleSize{CycleGAN} & \tableTitleSize{Ours} \\

\end{tabular}%
\begin{tabular}{c c c}

        \includegraphics[tSize,frame,]{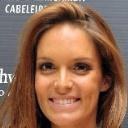} &
        \includegraphics[tSize,frame,]{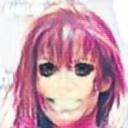} &
        \includegraphics[tSize,frame,]{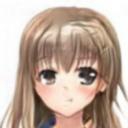}\\
        
        \includegraphics[tSize,frame,]{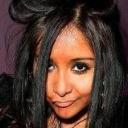} &
        \includegraphics[tSize,frame,]{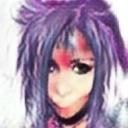} &
        \includegraphics[tSize,frame,]{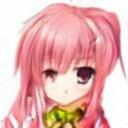}\\
        
        \includegraphics[tSize,frame,]{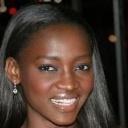} &
        \includegraphics[tSize,frame,]{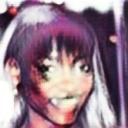} &
        \includegraphics[tSize,frame,]{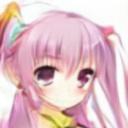}\\
        
        \tableTitleSize{Input} & \tableTitleSize{CycleGAN} & \tableTitleSize{Ours} \\

\end{tabular}%
\begin{tabular}{c c c}

        \includegraphics[tSize,frame,]{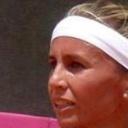} &
        \includegraphics[tSize,frame,]{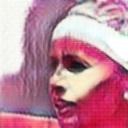} &
        \includegraphics[tSize,frame,]{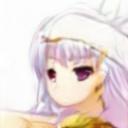}\\
        
        \includegraphics[tSize,frame,]{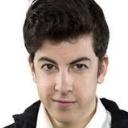} &
        \includegraphics[tSize,frame,]{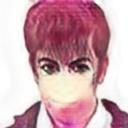} &
        \includegraphics[tSize,frame,]{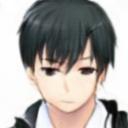}\\
        
        \includegraphics[tSize,frame,]{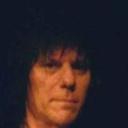} &
        \includegraphics[tSize,frame,]{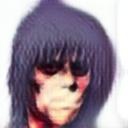} &
        \includegraphics[tSize,frame,]{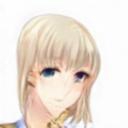}\\
        
        \tableTitleSize{Input} & \tableTitleSize{CycleGAN} & \tableTitleSize{Ours} \\

\end{tabular}%
\label{tab:getchu}
\end{table}

\begin{table}
    \makeatletter
    \define@key{Gin}{tSize}[true]{%
        \edef\@tempa{{Gin}{width=.12\linewidth, keepaspectratio}}%
        \expandafter\setkeys\@tempa
    }
    \makeatother
    \caption{Bitmoji to Human: Our network mode collapses down to a single image (right-hand side), even though our reconstructions for these samples are almost identical to the input image. This implies that our network is able to encode the entire database into this single sample using almost imperceptible differences in the image \cite{chu2017cyclesteg}. The failure is likely caused by the simplicity of the bitmoji domain, e.g., uniform skin color.}
    \setlength{\tabcolsep}{0.00em}
    \renewcommand{\arraystretch}{0.01}%
    \providecommand\aport{}
    \newcommand{\tableTitleSize}[1]{\tiny{#1}}
    
    \centering
\begin{tabular}{c c c c}
        \tableTitleSize{Input} & \tableTitleSize{CycleGAN} & \tableTitleSize{Ours} & \tableTitleSize{Reconstr.} \\

        \includegraphics[tSize,frame,]{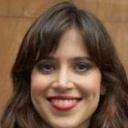} &
        \includegraphics[tSize,frame,, ]{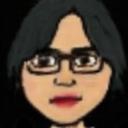} &
        \includegraphics[tSize,frame,]{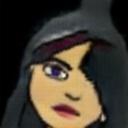} &
        \includegraphics[tSize,frame,]{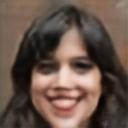}\\
        
        \includegraphics[tSize,frame,]{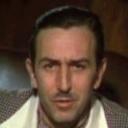} &
        \includegraphics[tSize,frame,, ]{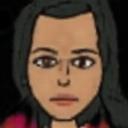} &
        \includegraphics[tSize,frame,]{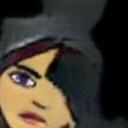} &   
        \includegraphics[tSize,frame,]{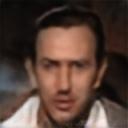}\\
        
        \includegraphics[tSize,frame,]{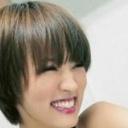} &
        \includegraphics[tSize,frame,, ]{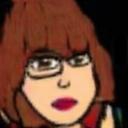} &
        \includegraphics[tSize,frame,]{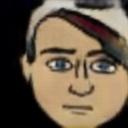} &         
        \includegraphics[tSize,frame,]{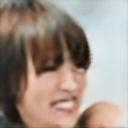}\\

        \tableTitleSize{Input} & \tableTitleSize{CycleGAN} & \tableTitleSize{Ours} & \tableTitleSize{Reconstr.} \\
 \end{tabular}%
\begin{tabular}{c c c c}

        \tableTitleSize{Input}  & \tableTitleSize{CycleGAN}&  \tableTitleSize{Ours} & \tableTitleSize{Reconstr.}\\

        \includegraphics[tSize,frame,]{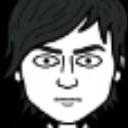} &
        \includegraphics[tSize,frame,, ]{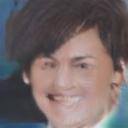} &
        \includegraphics[tSize,frame,]{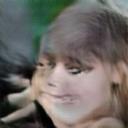} & 
        \includegraphics[tSize,frame,]{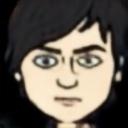}\\
        
        \includegraphics[tSize,frame,]{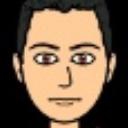} &
        \includegraphics[tSize,frame,, ]{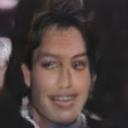} &
        \includegraphics[tSize,frame,]{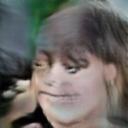}& 
        \includegraphics[tSize,frame,]{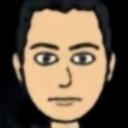}\\
        
        \includegraphics[tSize,frame,]{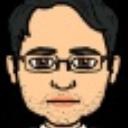} &
        \includegraphics[tSize,frame,, ]{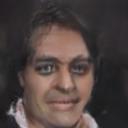} &
        \includegraphics[tSize,frame,]{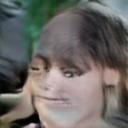} & 
        \includegraphics[tSize,frame,]{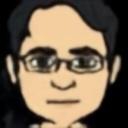} \\
    
        \tableTitleSize{Input} & \tableTitleSize{CycleGAN} & \tableTitleSize{Ours} & \tableTitleSize{Reconstr.} \\
        
\end{tabular}
\label{tab:bitmoji}
\end{table}

\end{document}